\documentclass[10pt,journal,compsoc]{IEEEtran}

\usepackage{hyperref}       
\hypersetup{colorlinks=true,urlcolor=black}
\usepackage{booktabs} 

\usepackage{capt-of}
\usepackage{xspace}
\usepackage{enumitem}
\usepackage{bbding}
\usepackage{amsmath,amssymb,amsbsy,amsfonts,dsfont,pifont,bm,bbm,mathrsfs,mathtools,nicefrac}
\usepackage{algorithm,algpseudocode,listings}
\usepackage{booktabs,multirow,adjustbox,diagbox,threeparttable,tabularray,colortbl}

\usepackage{indentfirst} 

\newcommand{\method}{\texttt{ReplaceAnyone}\xspace}

\definecolor{Gray}{gray}{0.94}
\definecolor{liGray}{gray}{0.5}
\definecolor{LightCyan}{rgb}{0.88,1,1}

\usepackage{colortbl} 
\usepackage{xcolor}

\usepackage[capitalize]{cleveref}  
\crefname{section}{Sec.}{Secs.}
\Crefname{section}{Section}{Sections}
\crefname{table}{Tab.}{Tabs.}
\Crefname{table}{Table}{Tables}
\crefname{figure}{Fig.}{Figs.}
\Crefname{figure}{Figure}{Figures}
\crefname{equation}{Eq.}{Eqs.}
\Crefname{equation}{Equation}{Equations}
\usepackage{xspace}
\newcommand{\ie}{{\emph{i.e.}}\xspace}

\newcommand{\eg}{{\emph{e.g.}}\xspace}

\newlength\savewidth\newcommand\shline{\noalign{\global\savewidth\arrayrulewidth
  \global\arrayrulewidth 1pt}\hline\noalign{\global\arrayrulewidth\savewidth}}

\newcommand\wx[1]{{\color{black}#1}}

\usepackage{ragged2e}

\ifCLASSINFOpdf
\else
\fi

\begin{document}

\title{Replace Anyone in Videos}

\author{Xiang~Wang, 
    Shiwei~Zhang,
    Haonan~Qiu,
    Ruihang~Chu,
    Zekun~Li,
    Yingya~Zhang,
    Changxin~Gao, 
     Yuehuan~Wang, 
     Chunhua~Shen,
     and Nong~Sang

\IEEEcompsocitemizethanks{
\IEEEcompsocthanksitem X. Wang, C. Gao, Y. Wang and N. Sang are with the Key Laboratory of Ministry of Education for Image Processing and Intelligent Control, School of Artificial
Intelligence and Automation, Huazhong University of Science and Technology, Wuhan, 430074, China. E-mail: (wxiang, cgao, yuehwang, nsang)@hust.edu.cn.
\IEEEcompsocthanksitem S. Zhang, R. Chu and Y. Zhang are with Alibaba Group, Hangzhou, 310052, China. E-mail: (zhangjin.zsw, churuihang.crh, yingya.zyy)@alibaba-inc.com.
\IEEEcompsocthanksitem H. Qiu is with Nanyang Technological University, 639798, Singapore. E-mail: qhnmoon@gmail.com.
\IEEEcompsocthanksitem Z. Li is with Nanjing University, Nanjing, 210008, China. E-mail: lizekun@smail.nju.edu.cn.
\IEEEcompsocthanksitem C. Shen is with Zhejiang University, Hangzhou, 310058, China. E-mail: chunhuashen@zju.edu.cn.
}
}


\markboth{Journal of \LaTeX\ Class Files,~Vol.~14, No.~8, August~2015}%
{Shell \MakeLowercase{\textit{et al.}}: Bare Demo of IEEEtran.cls for Computer Society Journals}


\IEEEtitleabstractindextext{
\begin{abstract}
\justifying
The field of controllable human-centric video generation has witnessed remarkable progress, particularly with the advent of diffusion models.
However, achieving precise and localized control over human motion in videos, such as replacing or inserting individuals while preserving desired motion patterns, still remains a formidable challenge.
In this work, we present the \method framework, which focuses on localized human replacement and insertion featuring intricate backgrounds.
Specifically, we formulate this task as an image-conditioned video inpainting paradigm with pose guidance, utilizing a unified end-to-end video diffusion architecture that facilitates image-conditioned video inpainting within masked regions.
To prevent shape leakage and enable granular local control, we introduce diverse mask forms involving both regular and irregular shapes.
Furthermore, we implement an enriched visual guidance mechanism to enhance appearance alignment, a hybrid inpainting encoder to further preserve the detailed background information in the masked video, and a two-phase optimization methodology to simplify the training difficulty.
\method enables seamless replacement or insertion of characters while maintaining the desired pose motion and reference appearance within a single framework.
%
Extensive experimental results demonstrate the effectiveness of our method in generating realistic and coherent video content.
The proposed \method can be seamlessly applied not only to traditional 3D-UNet base models but also to DiT-based video models such as Wan2.1.
The code will be available at \url{https://github.com/ali-vilab/UniAnimate-DiT}.

\end{abstract}
}
\maketitle

\IEEEpeerreviewmaketitle

\begin{figure*}[t]  
\centering
\includegraphics[width=0.99\textwidth]{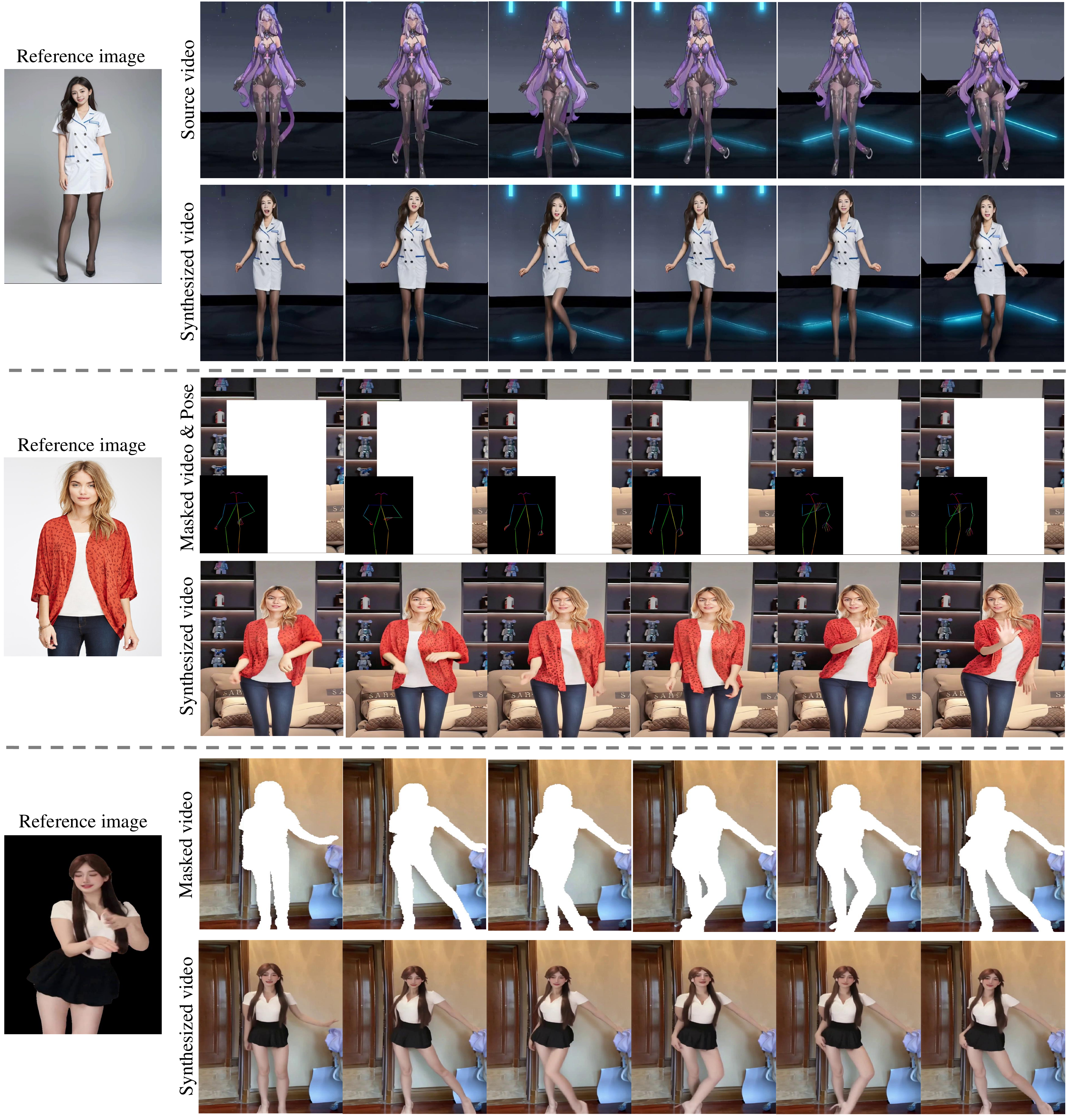}
\vspace{-3mm}
\caption{
\textbf{Video demo examples} synthesized by the proposed \method.
        Our \method enables character replacement or insertion in a source video with dynamic backgrounds using a reference image, preserving both the desired pose motion and reference appearance.
}
\label{first_figure}
\end{figure*}

\section{Introduction}
\label{sec:intro}


%

\IEEEPARstart{I}{n} recent years,
the field of controllable human-centric video generation~\cite{Animateanyone,ma2023dreamtalk,magicanimate,wang2024unianimate,magicdance,jiang2022text2human,karras2023dreampose,chai2020crowdgan,hong2023dagan++,zhu2023human,disco} has witnessed significant advancements, driven by the rapid development of generative models, particularly diffusion models~\cite{DDIM,DDPM,stablediffusion,Dalle2}. These models have enabled the creation of highly realistic and temporally coherent human videos, opening up new possibilities in various domains such as film production, virtual reality, and interactive media~\cite{kong2024hunyuanvideo,song2024directorllm}. 
However, achieving precise and localized control over human motion in videos, such as replacing or inserting individuals while preserving desired motion patterns and maintaining the authenticity of the scene, remains a formidable challenge. 

Existing methods in human-centric video generation~\cite{Animateanyone,wang2024unianimate,magicanimate,magicdance} primarily focus on leveraging human pose sequences and reference images to synthesize lifelike videos.
Notable works such as Animate Anyone~\cite{Animateanyone} and UniAnimate~\cite{wang2024unianimate} employ video diffusion models to capture spatial-temporal dependencies, enabling the generation of photorealistic human dance videos that align with given pose sequences and follow the appearance guidance of reference images. However, these methods are limited to synthesizing simple static backgrounds, which fail to precisely control the synthesis of complex, dynamic backgrounds.
This limitation significantly restricts their applicability in scenarios where dynamic scene interactions are required.
Another line of methods, \eg, TokenFlow~\cite{geyer2023tokenflow} and AnyV2V~\cite{ku2024anyv2v}, aims to edit source video content for global modifications. Despite significant progress, existing video editing methods still struggle to control local human motion and appearance effectively within dynamic scene videos~\cite{sun2024diffusion,tu2024motioneditor,tu2024motionfollower}. For instance, \emph{how to replace or insert a character while maintaining the desired appearance and pose
motion?}
Here, the challenges lie in seamlessly integrating the character in a specified location, ensuring it matches both the visual characteristics of the reference image and the desired motion indicated by the pose sequence, while leaving the untouched background undisturbed.

An intuitive solution~\cite{qin2023dancing} to character replacement \& insertion involves two steps.
It first fills in the masked video using a video inpainting model~\cite{li2022towards,gu2023flow,zhang2024avid,lee2024video}, and then pastes the animated character synthesized by an avatar animation method~\cite{wang2024unianimate,Animateanyone} back into the inpainted region. 
Yet, this paradigm requires multiple models to collaborate, which often leads to error accumulation and visible boundary mismatches. Besides, coordinating multiple models significantly increases computational overhead and model complexity, making it less efficient and hard to scale.

In this work, we tackle these challenges by introducing a novel, unified end-to-end framework denoted as \method.
It aims to achieve localized human replacement and insertion while preserving the authenticity of the scene. 
\wx{
\method aims to solve the following challenges:
\textbf{1) Shape leakage}: when using precise character masks to define the regions to be replaced or inserted and train the model, the obtained model may overfit to the shape information in the mask~\cite{wang2023imagen,zheng2022image}, leading to visible artifacts and inconsistencies in the generated video during the inference stage;
\textbf{2) Appearance alignment}: ensuring that the generated character matches the visual characteristics of the reference image while maintaining coherence with the background is crucial. 
\textbf{3) Background preservation}: preserving the detailed background information while filling in the masked regions is essential for seamless integration. 
\textbf{4) Optimization complexity}: training a unified model for both character insertion and inpainting presents significant optimization challenges. Balancing the objectives of generating realistic human motion and preserving background coherence is non-trivial.
}

To alleviate the above challenges, we present the \method framework, consisting of some tailored components.
From a data perspective, we design diverse mask forms involving both regular and irregular shapes to prevent potential leaks of shape information and accommodate different granularities of localized controls in the inference stage. 
This approach allows adaptability to different shape configurations, greatly enhancing operational flexibility. 
For reference character preservation, we propose an enriched visual guidance mechanism, which extracts semantic, shape, pose, and detail information from the reference image as rich guidance, so as to enhance the appearance alignment between the generated video and the reference character.
\wx{
Furthermore,
we empirically observed that if the common VAE encoder is used to encode masked video, the details of the background are not maintained well. We thus propose a hybrid inpainting encoder, which alleviates the above problem by adding an extra learnable light-weight
inpainting encoder to incorporate complementary details.
}
In addition, we advocate for a two-phase training strategy to reduce training complexity. The first phase trains an image-conditioned pose-driven video generation model, followed by a joint training of the video inpainting task on the masked regions.
In this way, our integrated approach, \method, achieves remarkable efficacy and sets a new paradigm  in localized human motion manipulation, as displayed in ~\cref{first_figure}.
Experimental results based on 3D-UNet and DiT models extensively reveal the effectiveness of \method in synthesizing high-fidelity consistent videos whose filled content can be seamlessly integrated with the dynamic backgrounds. 

\wx{
In summary, the main contributions of this work are as follows:
\begin{itemize}
\setlength{\parskip}{0pt} \setlength{\itemsep}{0pt plus 1pt}
    \item We present a novel \method framework for character replacement \& insertion and formulate this
task as an image-conditioned video inpainting paradigm with pose guidance, utilizing a unified end-to-end video diffusion architecture
that facilitates image-conditioned video inpainting within masked regions. 
    \item More comprehensively, we present diverse mask forms to prevent shape leakage, an enriched visual guidance mechanism to enhance appearance alignment, and  hybrid inpainting encoder to further preserve the detailed background information
in the masked video. 
    \item We advocate for a two-phase training strategy to reduce training complexity, which trains an image-conditioned pose-driven video generation model, followed by a joint training of the video inpainting task on the masked regions. 
    \item Both qualitative and quantitative experimental results  demonstrate the effectiveness of the proposed \method, showing that \method enables synthesizing
high-fidelity consistent videos whose filled content can be
seamlessly integrated with the dynamic backgrounds.
\end{itemize}
}

\section{Related Work}
\label{sec:related_work}

\begin{figure*}[t]
{
        \centering
    \includegraphics[width=0.999\textwidth]{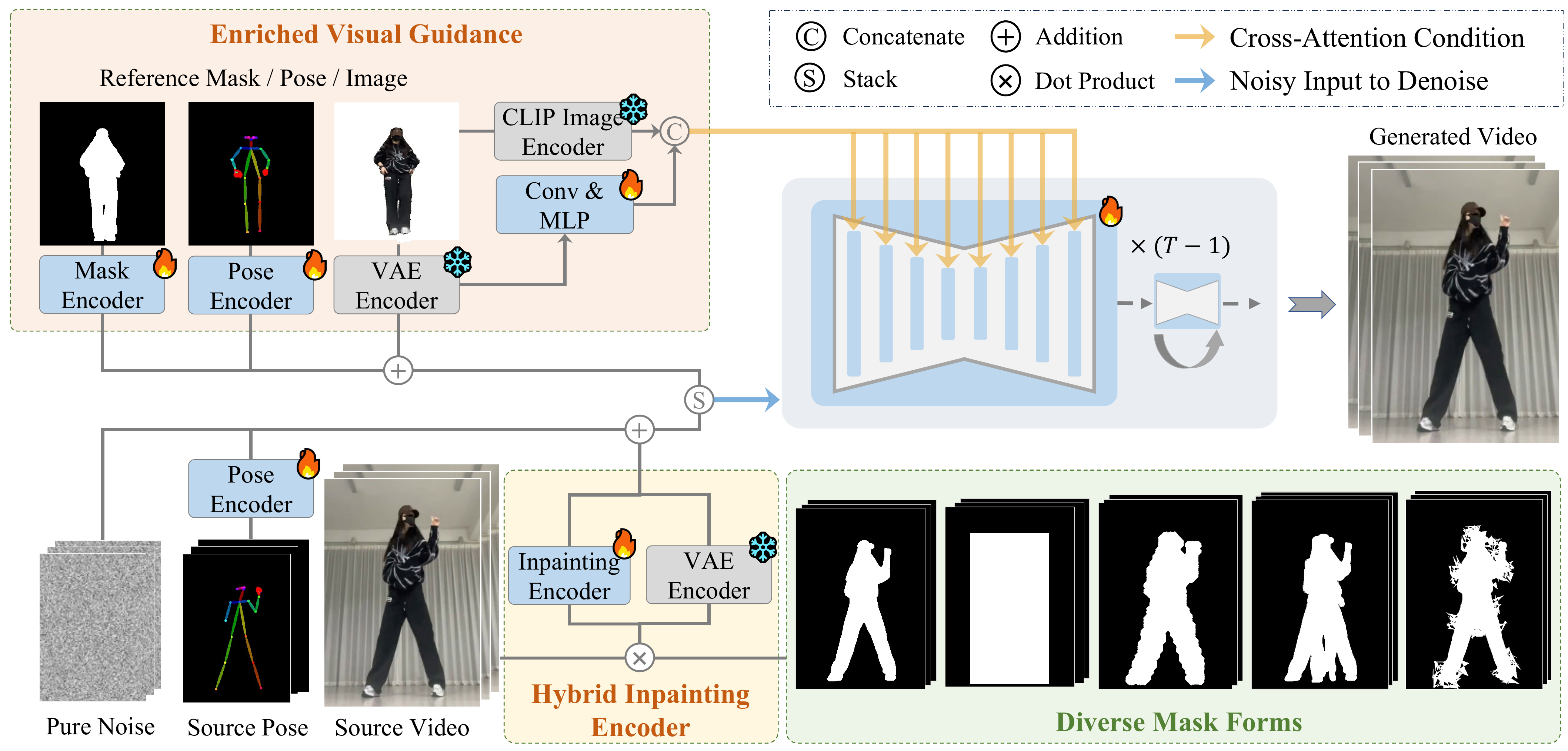}
    \vspace{-6mm}
    \caption{
        \textbf{Overall framework of \method
    }. 
    We use a unified video diffusion model to perform image-conditioned pose-driven video generation and video inpainting tasks simultaneously. In order to encode reference image information comprehensively, we design an enriched visual guidance mechanism to extract mask, pose and segmented image features respectively. Moreover, a variety of mask forms are designed to prevent the leakage of segmentation shape information and facilitate the fine-grained control.
    To preserve the details in the masked video, we design a hybrid inpainting encoder, which consists of a learnable inpainting encoder and a VAE encoder.
    Masked encoder, pose encoder, and inpainting encoder have similar structures, consisting of several learnable layers of downsampled convolutions to reduce computational complexity.
    %
    }
    \label{Network}
    }
\end{figure*}

\noindent \textbf{Human image animation.}
Given a reference character image and a desired pose sequence, this task aims to generate visually appealing and temporally consistent videos that adhere to the input conditions ~\cite{Animateanyone,wang2024unianimate,magicanimate,magicdance,disco,ma2024follow,lei2024comprehensive,huang2024magicfight,zhong2025posecrafter,peng2024controlnext,champ,MRAA,lei2024comprehensive,tan2024animate,qiu2024moviecharacter,tu2024stableanimator,meng2024echomimicv2,hu2025animate,zhang2024mimicmotion,men2024mimo,chang2025x,li2024dispose}.
Typically,
MagicAnimate~\cite{magicanimate} illustrates the potential of video diffusion models in generating motion-consistent videos driven by human pose sequences and designs a reference network to ensure appearance alignment between the reference image and the synthesized video.
Animate Anyone~\cite{Animateanyone} presents a ReferenceNet to merge detailed reference features via spatial attention to preserve the appearance consistency of the intricate reference image and designs an efficient pose guider to learn the movements of the character.
Champ~\cite{champ} leverages many rendered inputs, including depth images, normal maps, and semantic maps, to collectively improve shape alignment and motion guidance in the human image animation task.
UniAnimate~\cite{wang2024unianimate} proposes a unified video diffusion model to handle both appearance and motion alignment and incorporates the information of the reference pose to improve body correspondences.  
Different from these methods, which excel in synthesizing motion from a single reference image and can not be used to precisely preserve/generate desired scene dynamics, our approach attempts to locally control the movement and appearance of the characters and maintain the moving backgrounds in a dynamic video.

\vspace{1mm}
\noindent \textbf{Video editing.} 
With the rapid development of text-to-video models, \eg, ModelScopeT2V~\cite{modelscopet2v}, AnimateDiff~\cite{guo2023animatediff}, CogVideoX~\cite{yang2024cogvideox}, and HunyuanVideo~\cite{kong2024hunyuanvideo}, many researchers start to apply them into downstream tasks.  
Recently, there have been remarkable advances in video editing tasks~\cite{ouyang2024i2vedit,mou2024revideo,gu2024via,zhang2023controlvideo,tu2024motioneditor,zhao2025motiondirector,zhuo2024fast,ceylan2023pix2video,zhao2023controlvideo} with various guidance signals, including depth maps~\cite{Gen-1,videocomposer,zhang2023controlvideo}, optical flow~\cite{cong2023flatten}, human poses~\cite{pang2023dpe,tu2024motioneditor,zhong2024deco}, neural layered atlas~\cite{kasten2021layered,chai2023stablevideo}, etc. 
Representative works such as Gen-1~\cite{Gen-1} and Make-Your-Video~\cite{xing2023make} leverage depth maps to maintain the structural information of source videos and can synthesize video style content with the textual and image conditions.
VideoComposer~\cite{videocomposer} introduces a compositional video generation paradigm that allows users to flexibly compose a customized video with various input conditions, including textual conditions, spatial
conditions, and temporal conditions.
Pix2Video~\cite{ceylan2023pix2video} tries to leverage a pre-trained 2D
image model with structure guidance to edit an anchor frame and
progressively propagate the changing features to subsequent frames through injecting
self-attention features.
TokenFlow~\cite{geyer2023tokenflow} attempts to edit some  tokens of key frames and proposes to propagate the edited tokens across the video through pre-computed correspondences.
AnyV2V~\cite{ku2024anyv2v} first employs an off-the-shelf image editing network  to manipulate the first frame and then applies  
 DDIM inversion and feature replacement techniques to edit video content.
%
Nevertheless,
existing methods primally  focus on global content editing or local appearance manipulations and still
face challenges in effectively controlling local human
motion following a given character's identity  within dynamic scene videos.

\vspace{1mm}
\noindent \textbf{Video inpainting.}
Erasing a part of the video, this task requires filling the video content and keeping it in harmony with the unerased parts~\cite{kim2019deep,zeng2020learning,zhou2023propainter}.
Previous works~\cite{xu2019deep,li2022towards,zhang2022flow} usually employ hybrid CNN and Transformer architectures to capture both short-range and long-range dependencies with the assistance of predicted optical flow completion for video inpainting.
Some recent methods~\cite{gu2023flow,lee2024video,zhang2024avid} started to apply diffusion models to iteratively fill in visual content with various conditions, such as textual guidance, sketch signals, etc.
These methods have achieved great success and are widely used in many downstream tasks~\cite{quan2024deep}.
Typical works such as FGT~\cite{zhang2022flow} and E$^2$FGVI~\cite{li2022towards} are techniques that are widely discussed and adopted.
FGT~\cite{zhang2022flow} designs a new convolution-based flow completion model to estimate completed flows from
the corrupted ones by leveraging the corresponding flow representation in a local
temporal window and uses a flow-guided transformer network to generate the contents for remaining blank regions.
E$^2$FGVI~\cite{li2022towards} presents a flow-guided end-to-end framework, which consists of
three learnable modules, namely, flow completion, feature
propagation, and content hallucination modules, and the three modules can be jointly optimized, yielding a
highly efficient and effective pipeline.
%
%
%
 In this paper, we focus on the more challenging conditional inpainting task that requires local control of human pose motion and  character appearance in the filling area.

\section{Method}

In this section, we introduce the proposed framework, \method, which enables localized manipulation of  human appearance and motion in dynamic video scenes.
Firstly, we will provide a brief overview of  video diffusion models.
Subsequently, the algorithmic details of the proposed method will be presented.
%
The overall pipeline of \method is illustrated in \cref{Network}.

\subsection{Preliminaries of diffusion models}
%
The development of diffusion models~\cite{an2023latent,he2022latent,controlnet,stablediffusion} has promoted rapid progress in the field of video generation~\cite{zhou2022magicvideo,modelscopet2v,tft2v,VideoLDM,yuan2023instructvideo,ma2023dreamtalk,wei2023dreamvideo,zhang2023i2vgen,qing2023hierarchical,chen2023videocrafter1,imagenvideo,tune-a-video,make-a-video,cogvideo,wang2023videofactory,ceylan2023pix2video,Text2video-zero,lee2024video,he2023scalecrafter}.
These models involve a forward diffusion stage and a reversed denoising process,  which first gradually adds random noise to the clean video latent $x_{0}$ and then learns to  iteratively denoise the noised latent representation until it converges to a visually coherent output.
The forward diffusion operation can be formulated as:
\begin{equation}
  q(x_{t}|x_{t-1}) = \mathcal{N}(x_{t};\sqrt{1-\beta_{t}}x_{t-1}, \beta_{t}I)
  \label{eq:1}
\end{equation}
where $t=1,...,T$ denotes timestep, and $\beta_{t} \in (0,1)$ represents the pre-defined noise schedule, controlling the noise strength at each step.  
To ensure that
 the final $x_{T}$ conforms to a random Gaussian distribution $\mathcal{N}(0,I)$, the $T$ is usually set to a large value, \eg, $T=1000$.
After the forward diffusion stage, the goal of the reversed denoising process is to iteratively estimate the noise of $x_{T}$:
\begin{equation}
p_{\theta }(x_{t-1}|x_{t}) = \mathcal{N}(x_{t-1};\mu_{\theta}(x_{t},t),{\textstyle \sum_{\theta}}(x_{t},t) )
  \label{eq:2}
\end{equation}
%
To achieve this goal, a denoising model  parameterized by $\theta$, denoted as $\hat{x}_{\theta}$, is employed to approximate the original data $x_{0}$.
In typical implementations, $\hat{x}_{\theta}$ is a 3D-UNet~\cite{VideoLDM,modelscopet2v} or a DiT-based model~\cite{wang2025wan,kong2024hunyuanvideo}, which is used to model spatio-temporal dependencies within videos.
%
%
The training objective is:
\begin{equation}
\mathcal{L}_{base} = \mathbb{E}_{\theta}[|| v - \hat{x}_{\theta}(x_{t},t,c) ||_{2}^{2}]
  \label{eq:3}
\end{equation}
where $c$ denotes the  conditional guidance such as reference image and pose sequence, and $v$ means the parameterized target. 
%

\subsection{ReplaceAnyone}

%
The objective of \method is to achieve precise and localized control over human appearance and motion, that is, to seamlessly replace or insert characters into videos while preserving desired motion patterns without disturbing unmasked regions.
To achieve this goal,
we attempt to integrate image-conditioned pose-driven video generation and masked video inpainting tasks into a unified framework.
We introduce several novel components, including diverse mask forms, an enriched visual guidance mechanism, a hybrid inpainting encoder, and a two-phase training strategy. These components collectively address the challenges of shape leakage, appearance alignment, background preservation, and optimization complexity.

\begin{figure}[t]
{
        \centering
    \includegraphics[width=0.49\textwidth]{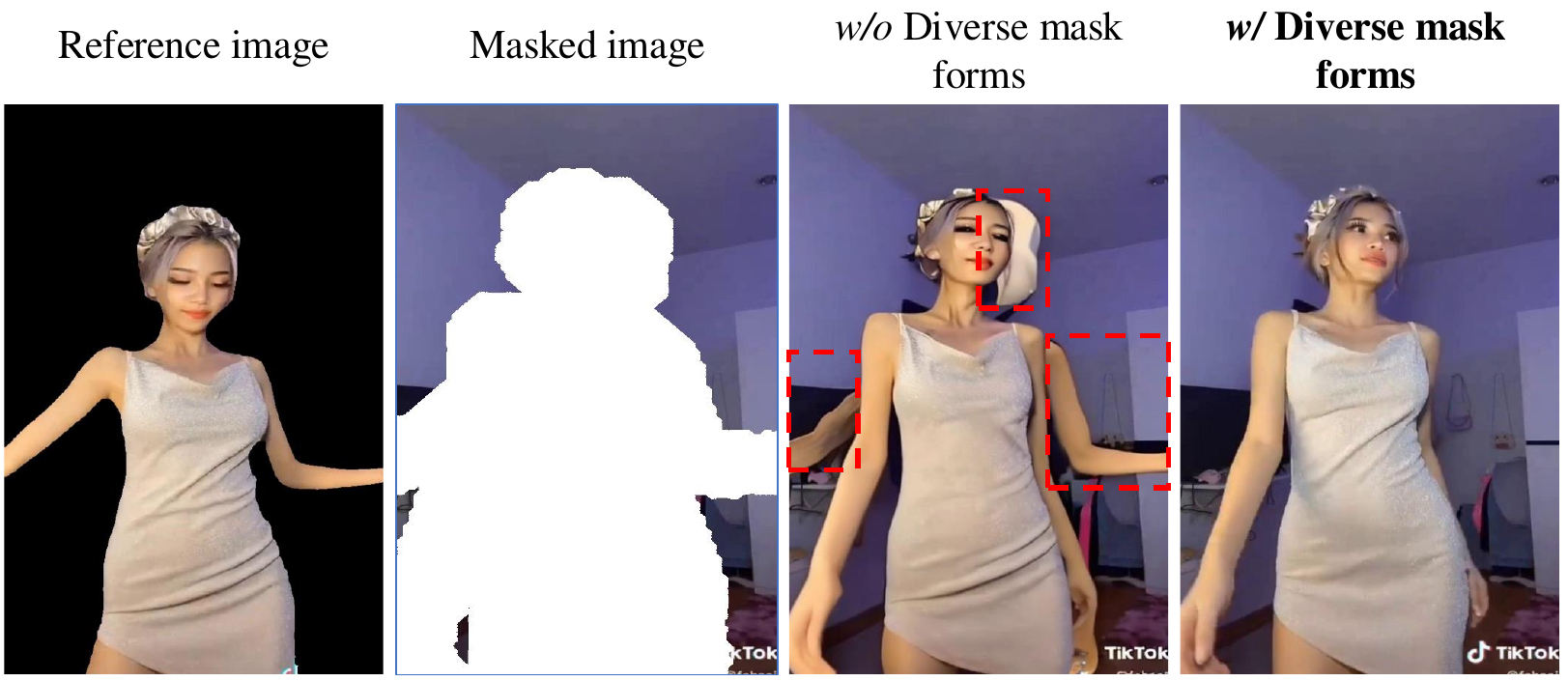}
    \vspace{-8mm}
    \caption{
    \wx{
        \textbf{Illustration of shape leakage}. If only the original character mask is used for training, the network will overfit to the information of the masked shape, resulting in obvious discordant parts during inference, such as ``four hands" or unrealistic padding.
        This problem can be significantly alleviated by introducing diverse mask forms.
        }
    }
    \label{08_shape_leakage}
    }
\end{figure}

\subsubsection{Diverse mask forms}
\vspace{1mm}
%
For inpainting, a direct approach is to accurately mask the area that needs to be modified and then fill it according to the input conditions. However, as displayed in~\cref{08_shape_leakage}, this method has the problem of shape leakage~\cite{wang2023imagen,zheng2022image}, since the mask contains the shape information of the content that needs to be manipulated, and the model will overfit this information, resulting in poor generalization ability.
To alleviate this issue and ensure flexible control, we develop multiple types of masks tailored for different granularities of local control. 
%
We categorize the masks into five forms:
%
1) Precise masks: these indicate the precise regions to be replaced and are useful for precise control over the insertion of the character.
2) Rectangular boundary masks, which represent the positional bounding box of the character in the video.
3) Inflated masks, which evolved from a precise mask through expansion operation.
4) Blended masks, obtained by mixing additional masks from the character in this video or other videos with the current mask, prevent shape leakage and enrich shape variety.
5) \wx{Edge destruction masks, randomly add some random shape masks to the edge of the human mask to break the edge shape information. In \cref{Network}, by applying this mask to the edge, it is no longer visible that the person is wearing a hat.}
These diverse masks serve as constraints during the inpainting process, enabling the integration of new content into these masked areas.

\wx{
These mask forms will be randomly sampled during training and dot-multiplied with the original video to get the masked video, and we need to learn to fill the masked video.
The diverse mask forms that we proposed have two advantages: 1) Prevent the leakage of shape information during the training process and help learn a highly generalized model; 2) Support multiple forms of masks during the inference process, including regular (such as boxes) and irregular (\eg, accurate masks), providing users with more flexibility.
}

\subsubsection{Enriched visual guidance}
\vspace{1mm}
%
Traditional animation methods~\cite{Animateanyone,wang2024unianimate,magicanimate} usually apply CLIP~\cite{CLIP} and Variational Autoencoder (VAE)~\cite{VAE} to extract reference image features, ignoring information such as shape and posture.
To encode comprehensive reference representation, we design an enriched visual guidance mechanism, which extracts several complementary information: mask features, pose features, and appearance VAE features derived from the segmented reference image. These features are then fused with noised video and fed into the video diffusion model.
Considering the limitations of CLIP features, which primarily capture global style information while neglecting local details, we enhance our feature extraction by incorporating local VAE features. The VAE image features are processed through multilayer perceptron (MLP) and convolution blocks to produce a downsampled representation. The obtained representation is then concatenated with the CLIP features, enriching the input for the video diffusion model through cross-attention mechanisms.
This enriched feature extraction enhances the appearance alignment between the generated video and the reference character, resulting in more realistic and coherent video synthesis.

\begin{figure}[t]
\vspace{+1mm}
{
        \centering
    \vspace{0.6mm}\includegraphics[width=0.49\textwidth]{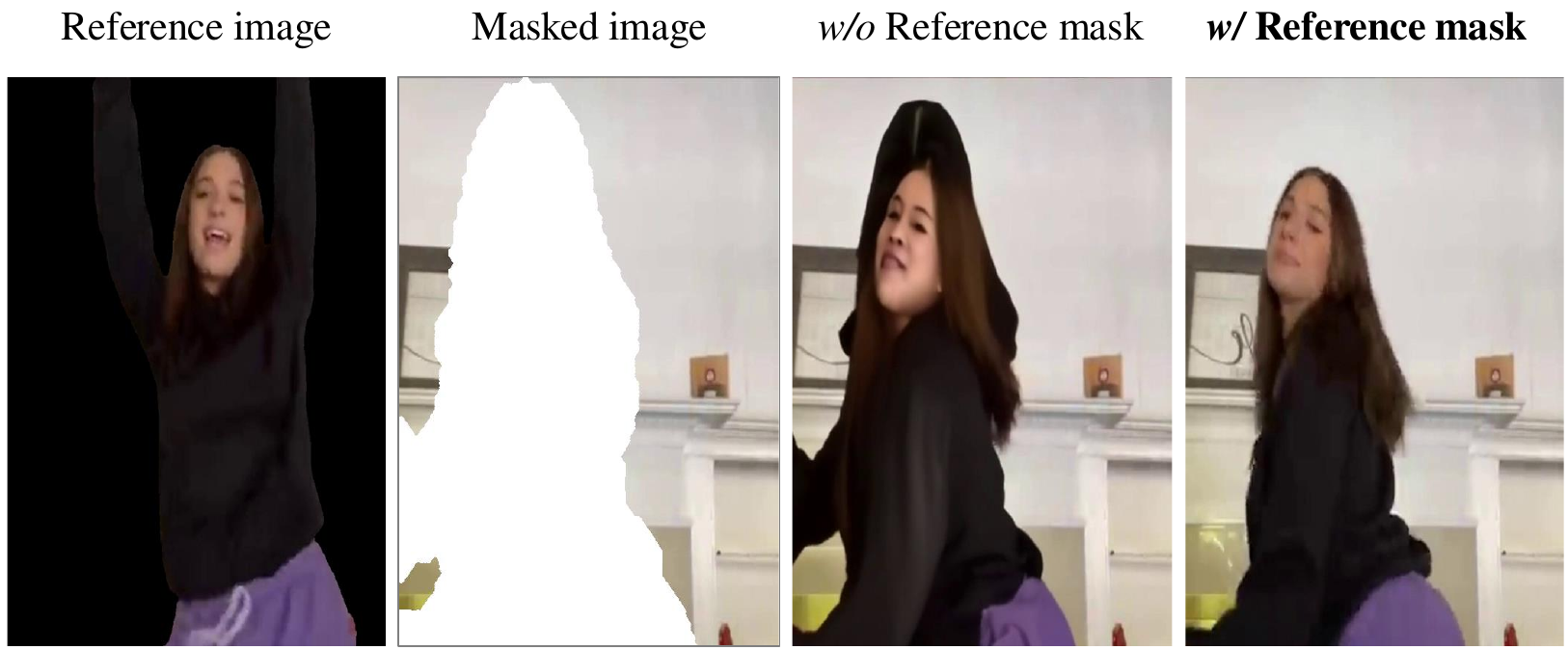}
    \vspace{-8mm}
    \caption{
    \wx{
        \textbf{Importance of incorporating reference mask}. If the reference mask is not introduced, the model may mistakenly regard the black background around the reference character as part of the character, resulting in unrealistic generation.
        }
    }
    \label{reference_mask}
    }
\end{figure}

In~\cref{reference_mask}, we visualize an example without incorporating reference mask.
From the results, we can observe that 
the model mistakenly regard the black background as part of the character, resulting in unreasonable generation.

\begin{figure}[t]
{
        \centering
    \vspace{0.6mm}\includegraphics[width=0.49\textwidth]{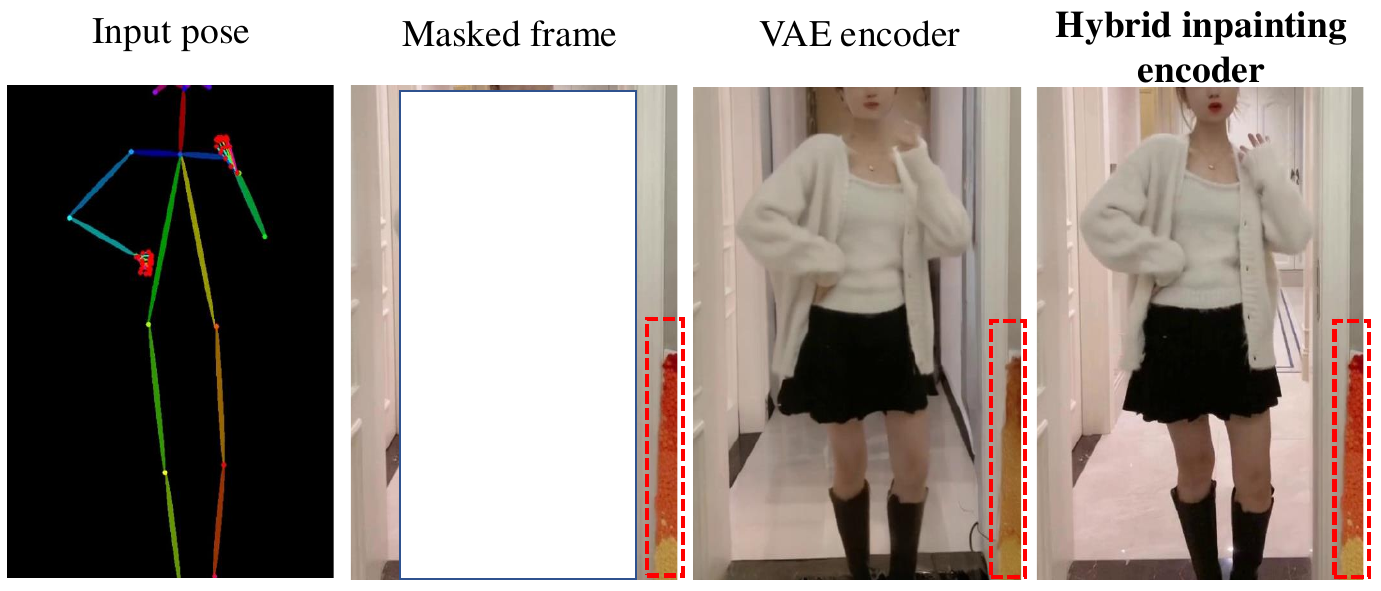}
    \vspace{-8mm}
    \caption{
    \wx{
        \textbf{Qualitative comparison} between VAE encoder and our hybrid inpainting encoder. The proposed hybrid inpainting encoder can utilize the powerful semantic encoding capability in VAE encoder and the detail preservation capability of learnable inpainting encoder to enhance the preservation effect for background information in the masked video frames.
        }
    }
    \label{07_compare_with_vae}
    }
\end{figure}

\begin{table}[t]
%
\footnotesize
\caption{
\textbf{Numerical MSE reconstruction error}.  We randomly sample 2,000 videos, feed complete video frames and masked frames into VAE for reconstruction, and calculate the reconstruction error of the same unmasked background area.
}
\vspace{-2mm}
\renewcommand{\arraystretch}{1.2}
\setlength\tabcolsep{5.5pt}
\centering
\begin{tabular}{l|cccc}
\shline
\hspace{-0.4mm}Frame type       & Complete frame   & Masked frame  \\ \shline
\hspace{-0.4mm}MSE reconstruction error ($\downarrow$)
 & 0.0012
  &   0.0023          \\

\shline
\end{tabular} 
\label{tab:mse_error}
\end{table}

\begin{figure}[t]
{
        \centering
    \vspace{0.6mm}\includegraphics[width=0.49\textwidth]{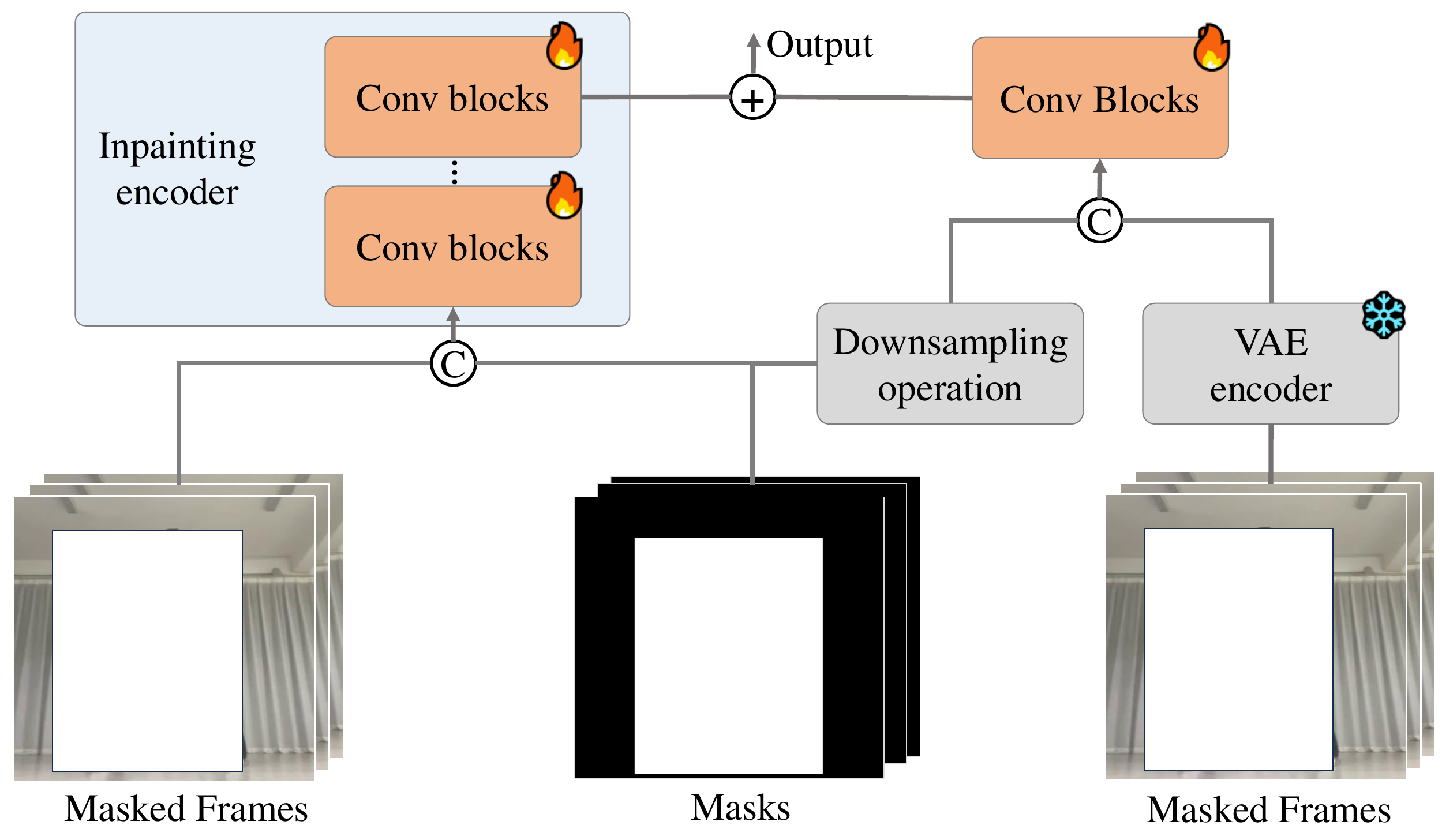}
    \vspace{-8mm}
    \caption{
    \wx{
        \textbf{Detailed illustration} of hybrid inpainting encoder. 
        Both masked frames and masks are used to extract features. 
        }
    }
    \label{Detail_hybrid_inpainting_encoder}
    }
\end{figure}

\subsubsection{Hybrid inpainting encoder}
\vspace{1mm}

\wx{
Encoding masked videos is crucial for preserving detailed background information. Traditional approaches~\cite{lee2024video} often use a pre-trained Variational Autoencoder (VAE)~\cite{VAE} to encode masked videos. However, we observed that the VAE, trained on complete images, may not perform optimally on masked inputs, leading to low local fidelity in the background, as shown in \cref{07_compare_with_vae}.
%
Since the filled parts are learned to coordinate with the background, poor compression and reconstruction of the background will cause the generated content in the filled parts to be affected by factors such as lighting and color.
In~\cref{tab:mse_error}, we calculate the average MSE reconstruction error of the unmasked background area based on 2,000 randomly sampled videos.
From the results, we can
observe that the reconstruction error of the masked frame is almost double that of the complete frame, indicating the loss of spatial information.
To alleviate this phenomenon, we propose a hybrid inpainting encoder that combines a pre-trained VAE encoder with a learnable lightweight inpainting encoder.
The detailed architecture of the hybrid inpainting encoder is displayed in~\cref{Detail_hybrid_inpainting_encoder}.
The proposed hybrid inpainting encoder leverages the powerful semantic encoding capability of the VAE while incorporating complementary details from the learnable encoder. This dual approach enhances the preservation of background information in the masked video frames, resulting in better video generation results that are highly consistent with the input masked video.
}

\subsubsection{two-phase  optimization strategy}
\vspace{1mm}

%
Since our framework needs to optimize two sub-tasks, this inevitably brings optimization and convergence challenges. To alleviate this problem, we explicitly split the optimization process into two steps.
%
In the first stage, we train a dedicated image-conditioned
pose-driven video generation model without the inpainting encoder like~\cite{wang2024unianimate} on human video datasets to establish a robust baseline capable of producing temporally coherent videos based on pose sequences and reference images.
In the second stage, we reuse the previously trained model and begin to jointly train with the inpainting of masked regions. Using the knowledge gained from the initial phase allows effective learning and reduces training difficulty, allowing the proposed \method to focus on generating consistent and harmonious results across both foreground and background layers.
This two-phase approach allows the model to first learn the fundamental skills of human image animation and then refine its capabilities for localized control and background preservation. 
%
Intuitively, the non-mask area is easy to learn. Instead, we should focus on the area to be filled. To this end, we propose a mask focused loss to focus on the area to be filled:
\begin{equation}
\mathcal{L} = \mathcal{L}_{base} + \alpha\cdot\mathbb{E}_{\theta}[|| mask*(v - \hat{x}_{\theta}(x_{t},t,c)) ||_{2}^{2}]
  \label{eq:4}
\end{equation}
where $\alpha$ is a hyperparameter, which is empirically set to 5. $mask$ represents the masked area.
This loss function ensures that the model focuses on generating high-quality content within the masked regions.

\begin{figure*}[t]
{
        \centering
    \includegraphics[width=1.0\textwidth]{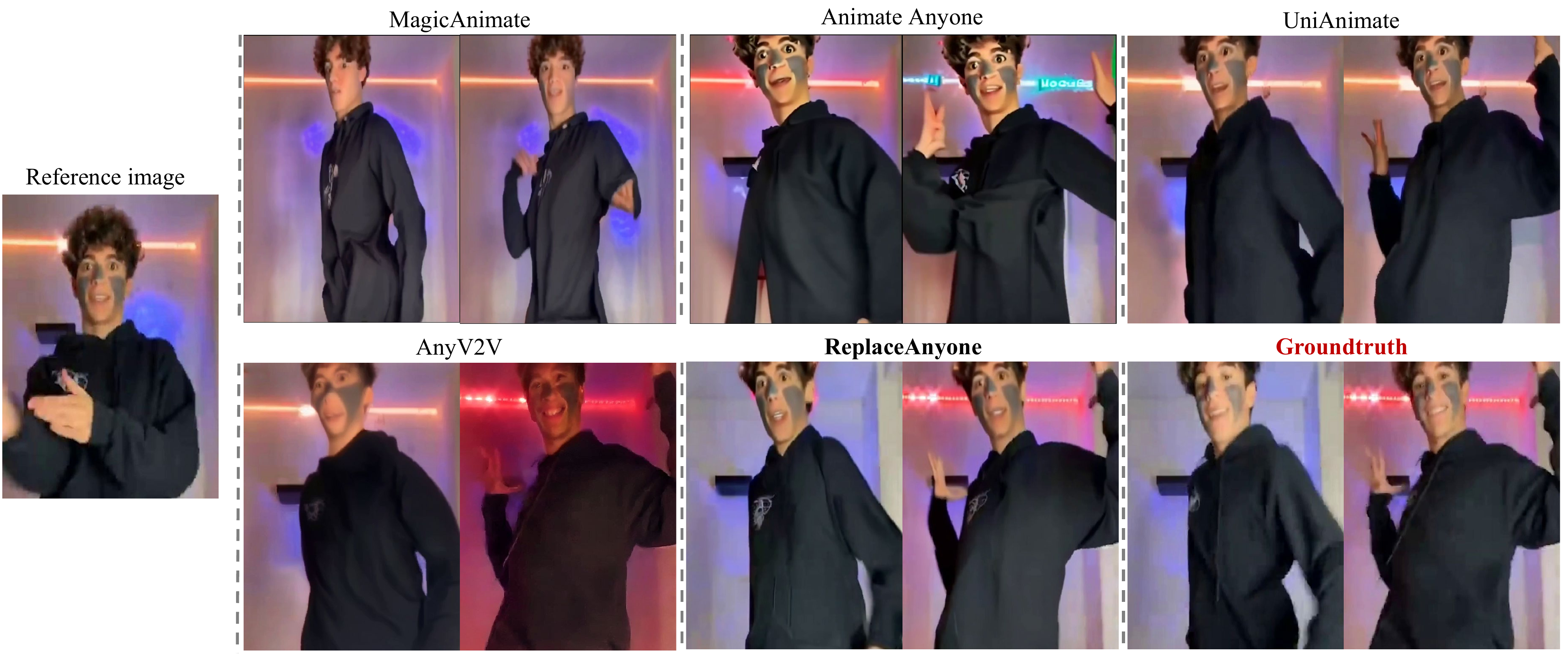}
    \vspace{-8mm}
    \caption{
    \wx{
        \textbf{Qualitative comparison} with existing human image animation methods~\cite{magicanimate,Animateanyone,wang2024unianimate} and video editing method AnyV2V~\cite{ku2024anyv2v}. 
        \wx{
        We find that existing methods fail to preserve
complex dynamic background while our method exhibits results very
close to the ground-truth.
}
        }
    }
    \label{06_compare_with_sota}
    }
     \vspace{-2mm}
\end{figure*}

\section{Experiments}

\begin{figure*}[t]
{
        \centering
    \includegraphics[width=1.0\textwidth]{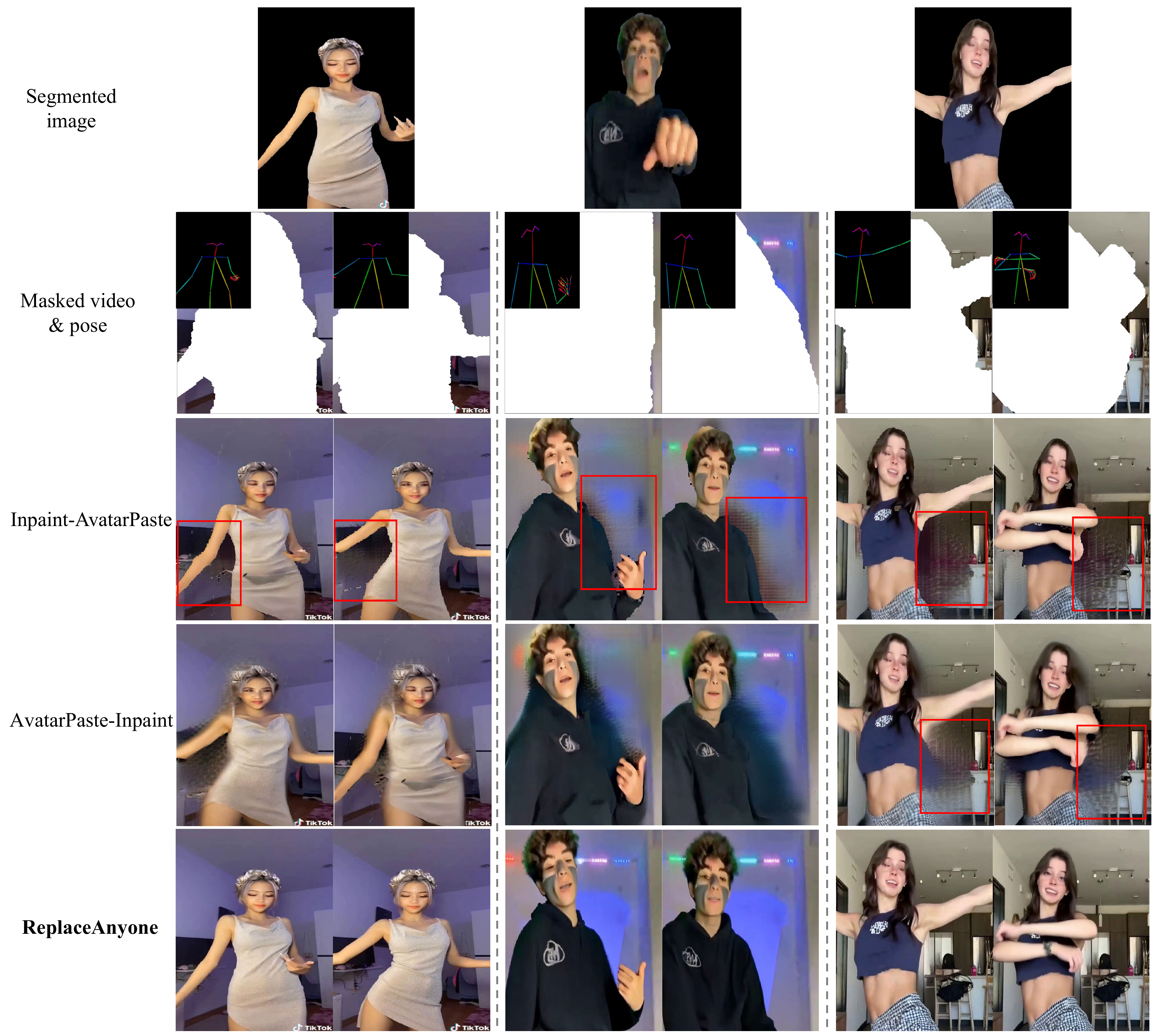}
    \vspace{-8mm}
    \caption{
        {\textbf{Qualitative comparison} between \method and baseline methods on the TikTok dataset}. 
        The baseline methods require multiple models to combine results, yielding noticeable blurred or mosaic-like artifacts.
        In contract, the videos generated by the proposed \method exhibit high-quality and temporal coherent results with smooth foreground-background connections. 
    }
    \label{04_comparison}
    }
    \vspace{-3mm}
\end{figure*}


In this section, we provide a detailed evaluation of the \method framework, focusing on its ability to perform localized human replacement and insertion in dynamic video scenes. We analyze the effectiveness of our proposed method through both quantitative metrics and qualitative visual comparisons.

%


\subsection{Experimental setup.} 

\noindent \textbf{Datasets and evaluation metrics.}
Due to the lack of publicly available large-scale human video datasets, we collected a custom dataset consisting of approximately 10,000 character dancing videos.
These videos feature diverse backgrounds, lighting conditions, and human motions, making the dataset suitable for training and evaluating our model.
Following prior works~\cite{Animateanyone,wang2024unianimate,magicanimate,champ}, we evaluate the performance using test sets from TikTok~\cite{tiktokdata} and UBC Fashion~\cite{UBCfashion} datasets, which include 10 and 100 videos, respectively. 
To quantitatively evaluate the performance of our method, we adopted several widely-used metrics: 1) PSNR (Peak Signal-to-Noise Ratio): PSNR is used to calculate the similarity between the generated video and the ground truth in terms of pixel-level accuracy; 2)
SSIM (Structural Similarity Index): Evaluates the structural similarity between the generated video and the ground truth, capturing both luminance and contrast; 3)
LPIPS (Learned Perceptual Image Patch Similarity): Provides a perceptual similarity score by comparing the generated video with the ground truth using a pre-trained neural network. 4)
FVD (Fréchet Video Distance)~\cite{unterthiner2018towards}: Measures the similarity between the distributions of generated and real videos, capturing both temporal and spatial coherence.
\wx{
In addition, to fully indicate the effectiveness of our method, human preference evaluation is also adopted.
30 video cases are randomly generated, and we ask three raters to vote on indicators, \ie, dynamic background preservation of original videos and temporal consistency of characters, with 10 representing the highest and 1 representing the lowest. 
}
\vspace{1mm}
\noindent \textbf{Implementation details.}
We extract pose sequences using the DwPose~\cite{DWpose} algorithm and obtain person segmentation masks with GroundedSAM~\cite{ren2024grounded}. 
For the first human image animation stage, we follow the optimization paradigm of UniAnimate~\cite{wang2024unianimate} to train our model. 
During the whole training process, we utilize Adam optimizer with a learning rate of 3e-5 to optimize the model.  All the experiments are conducted on 8 A100 NVIDIA GPUs (80G).
To ensure the efficacy of classifier-free guidance~\cite{ho2022classifierfreeguidance}, we apply a random dropout on input conditions with a ratio of 10\%.
\wx{
The training process of our \method consists of two stages.
For the first stage, we follow the training settings in UniAnimate~\cite{wang2024unianimate}
to train an image-conditioned pose-driven video generation model.
The mask encoder and the inpainting encoder are not involved during the training.
The architectures of the pose encoder and the 3D-UNet~\cite{VideoLDM} are the same as UniAnimate.
The mask encoder shares the same architecture as the pose encoder.
The hybrid inpainting encoder consists of a fixed VAE encoder and a small learnable head, which has a network structure similar to that of pose encoder.
During the second stage, we train the entire network end-to-end to learn to fill in the erased portion.
In inference,  the 50-step DDIM~\cite{DDIM} scheduler is utilized to synthesize videos.
}

\vspace{1mm}
\noindent \textbf{Baselines.}
A typical baseline type is the common human image animation techniques, including MagicAnimate~\cite{magicanimate}, Animate Anyone~\cite{Animateanyone}, UniAnimate~\cite{wang2024unianimate}. In addition, we also treat the state-of-the-art video editing method AnyV2V~\cite{ku2024anyv2v} as a baseline to verify the effectiveness of the proposed method.
To further demonstrate the effectiveness of our end-to-end method, we construct two strong baselines of multiple models in series: i) ``Inpaint-AvatarPaste" indicates a two-step method, where the masked video is first filled by a video inpainting model~\cite{li2022towards}, and then the animated character synthesized by~\cite{wang2024unianimate} is pasted back into the filled video to blend it into the final video;
ii)
``AvatarPaste-Inpaint" means that the animated character generated by~\cite{wang2024unianimate} is first pasted into the masked video, and then a video inpainting model~\cite{li2022towards} is applied to fill in the remaining blank areas.
\wx{
Furthermore, in~\cref{Comparison_with_image-to-video_models}, we also compare our method with existing state-of-the-art image-to-video models such as CogVideoX-5B-I2V~\cite{yang2024cogvideox} and HunyuanVideo-I2V~\cite{kong2024hunyuanvideo} to demonstrate the strengths of our method.
}

\subsection{Quantitative and qualitative evaluation}

\noindent \textbf{Comparison with existing state-of-the-art methods.} 
In~\cref{06_compare_with_sota}, we compare our \method with existing state-of-the-art human image animation methods, \ie, MagicAnimate~\cite{magicanimate}, Animate Anyone~\cite{Animateanyone}, UniAnimate~\cite{wang2024unianimate} and video editing method AnyV2V~\cite{wang2024unianimate}.
\wx{
Existing human image animation methods~\cite{magicanimate,Animateanyone,wang2024unianimate} leverage  a reference character image and a desired pose sequence as input to generate videos that adhere to the input conditions. However, as indicated in ~\cref{06_compare_with_sota}, these methods tend to preserve the background in the reference image or change the background randomly, failing to precisely maintain complex dynamic background.
AnyV2V~\cite{wang2024unianimate} tries to manipulate the
first frame and then applies DDIM inversion and feature
replacement techniques to edit video content. Yet, as shown in ~\cref{06_compare_with_sota}, AnyV2V primally focuses on global content editing and still faces challenges
in effectively preserving character details (\eg, facial details) and following complex background within dynamic scene videos.
}
In contrast, 
our method exhibits results very close to the ground-truth. 
This shows that our method has significant advantages in the local manipulation of human motion.
The human evaluation results in~\cref{tab:human_evaluation} also verify that \method outperforms other methods in dynamic background preservation and temporal consistency of the reference character.

\begin{table}[t]
%
\footnotesize
\caption{
\textbf{Human preference evaluation}. 
We randomly generate 30 video cases and ask three raters to vote on indicators, \ie, dynamic background preservation of original videos and temporal consistency of characters, with 10 representing the highest and 1 representing the lowest. The average scores are shown.
}
\vspace{-2mm}
\renewcommand{\arraystretch}{1.2}
\setlength\tabcolsep{2.0pt}
\centering
\begin{tabular}{l|cccc}
\shline
\hspace{-0.4mm}Method       & Dynamic preservation   & Character consistency  \\ \shline
\hspace{-0.4mm}MagicAnimate~\cite{magicanimate} 
 & 4.3
  &   5.6          \\
\hspace{-0.4mm}Animate Anyone~\cite{Animateanyone} 
 & 4.5
  &     7.6      \\
\hspace{-0.4mm}UniAnimate~\cite{wang2024unianimate} 
 & 4.4
  &     8.2      \\
\hspace{-0.4mm}AnyV2V~\cite{ku2024anyv2v} 
 & 6.8
  &     7.1      \\
\hspace{-0.4mm}{\textbf{\method}}
 & \textbf{8.1}      & \textbf{8.4}             \\

\shline
\end{tabular} 
\label{tab:human_evaluation}
\end{table}

\begin{figure*}[t]
{
        \centering
    \includegraphics[width=0.99\textwidth]{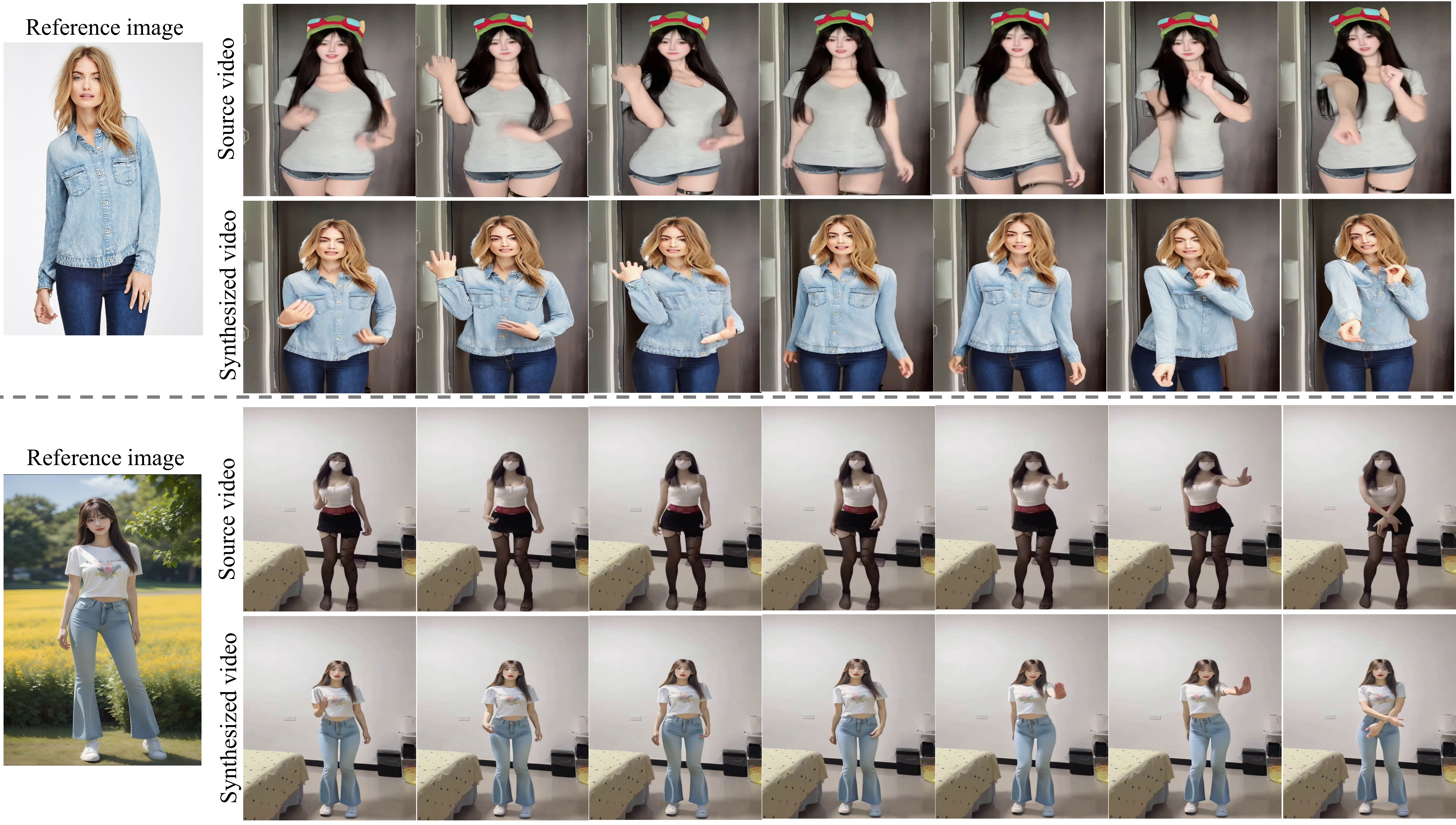}
    \vspace{-4mm}
    \caption{
        \textbf{{Qualitative results of character replacement}}. 
        The proposed \method can naturally and smoothly replace the person in the source video with the reference identity while maintaining the desired human poses.
    }
    \label{02_replacement}
    }
\end{figure*}

\begin{table}[t]
%
\caption{
{\textbf{Quantitative evaluation}  on  TikTok}. 
The ``Inpaint-AvatarPaste" method indicates a two-step paradigm, where the masked video is first filled by a video inpainting model~\cite{li2022towards}, and then the animated character synthesized by~\cite{wang2024unianimate} is pasted back into the filled video to blend it into the final video.
``AvatarPaste-Inpaint" means that the generated animated character is first pasted into the masked video, and then a video inpainting model is applied to fill in the other blank areas.
 }
 \vspace{-2mm}
\footnotesize
\renewcommand{\arraystretch}{1.2}
\setlength\tabcolsep{2.2pt}
\centering
\begin{tabular}{l|c|cccc}
\shline
\hspace{-0.8mm}Method       & End-to-end   & PSNR$\uparrow$ & SSIM$\uparrow$ & LPIPS$\downarrow$  & FVD$\downarrow$ \\ \shline
\hspace{-0.8mm}Inpaint-AvatarPaste~\cite{qin2023dancing} 
 & \XSolidBrush
  &   16.28      &      0.715       &             0.314                 &     404.86        \\
\hspace{-0.8mm}AvatarPaste-Inpaint 
 & \XSolidBrush
  &   16.14     &     0.713       &             0.307                &     374.93        \\
\hspace{-0.8mm}{\textbf{\method}}
 & \Checkmark      & \textbf{20.19}             & \textbf{0.820}            & \textbf{0.187}                                          &\textbf{183.66}  \\

\shline
\end{tabular} 
\label{tab:quantitative_TikTok}
\end{table}

\begin{table}[t]
%
\footnotesize
\caption{
\textbf{Quantitative evaluation}  on  the UBC Fashion dataset. ``Inpaint-AvatarPaste" indicates that the masked video is first filled by a video inpainting model~\cite{li2022towards}, and then the animated character synthesized by~\cite{wang2024unianimate} is pasted back into the filled video to blend it into the final video.
``AvatarPaste-Inpaint" means that the generated animated character is first pasted into the masked video, and then the video inpainting model is applied to fill in the other blank areas.}
\vspace{-2mm}
\renewcommand{\arraystretch}{1.2}
\setlength\tabcolsep{2.2pt}
\centering
\begin{tabular}{l|c|cccc}
\shline
\hspace{-0.8mm}Method       & End-to-end   & PSNR$\uparrow$ & SSIM$\uparrow$ & LPIPS$\downarrow$  & FVD$\downarrow$ \\ \shline
\hspace{-0.8mm}Inpaint-AvatarPaste~\cite{qin2023dancing} 
 & \XSolidBrush
  &   18.24      &       0.843       &                        0.129      &   546.28           \\
\hspace{-0.8mm}AvatarPaste-Inpaint 
 & \XSolidBrush
  &     21.30    &       0.883      &                 0.083           &      402.29        \\
\hspace{-0.8mm}{\textbf{\method}}
 & \Checkmark      & \textbf{22.86}             & \textbf{0.899}            & \textbf{0.069}                                          &\textbf{138.39}  \\

\shline
\end{tabular} 
\label{tab:quantitative_fashion}
\end{table}

\vspace{1mm}
\noindent \textbf{Comparison with strong baselines.} 
In addition, we further compare our method with two strong baselines.
\wx{
The first baseline is
``Inpaint-AvatarPaste", where the masked video is first filled by a video inpainting model~\cite{li2022towards}, and then the animated character synthesized by~\cite{wang2024unianimate} is pasted back into the filled video to blend it into the final video.
The second baseline is
``AvatarPaste-Inpaint", which means that the animated character generated by~\cite{wang2024unianimate} is first pasted into the masked video, and then a video inpainting model~\cite{li2022towards} is applied to fill in the remaining blank areas.
}
These baselines use multiple models in series to achieve local motion manipulation.
We conduct a qualitative comparison in~\cref{04_comparison}, and the visualizations indicate that the baseline methods suffer from serious mosaic-like artifacts at the boundary regions between character and background derived from disharmony painting. 
In contrast, our method displays smooth content, verifying   the effectiveness of \method and showcasing the advantages of the end-to-end approach in mitigating error accumulation.
We also quantitatively evaluate \method using standard metrics such as PSNR, SSIM, LPIPS, and FVD on the TikTok and UBC Fashion datasets. The results, as shown in~\cref{tab:quantitative_TikTok} and~\cref{tab:quantitative_fashion}, exhibit that our method reaches best performance in all metrics,  demonstrating the superiority of  \method over baseline methods in terms of both visual fidelity and video quality.
This suggests that our method effectively preserves complex dynamic backgrounds while seamlessly integrating characters into videos.

\vspace{0.5mm}
\noindent \textbf{Character replacement.}
\wx{
Character replacement involves precisely replacing a person in a source video with a different character while maintaining the desired pose and appearance. This task is challenging because it requires the model to not only generate a new character that matches the reference image but also ensure that the character's motion aligns with the given pose sequence and blends harmoniously with the background.
}
For the character replacement task, we demonstrate the effectiveness of our framework by executing tests on individual video segments where we replace characters in a variety of scenarios, as displayed in~\cref{02_replacement}. The smooth and harmonious video generation results indicate that \method accurately captures reference appearance and movement, preserving the contextual integrity of the scene.
%


%
%

%

\begin{figure*}[!htbp]
{
        \centering
    \includegraphics[width=0.99\textwidth]{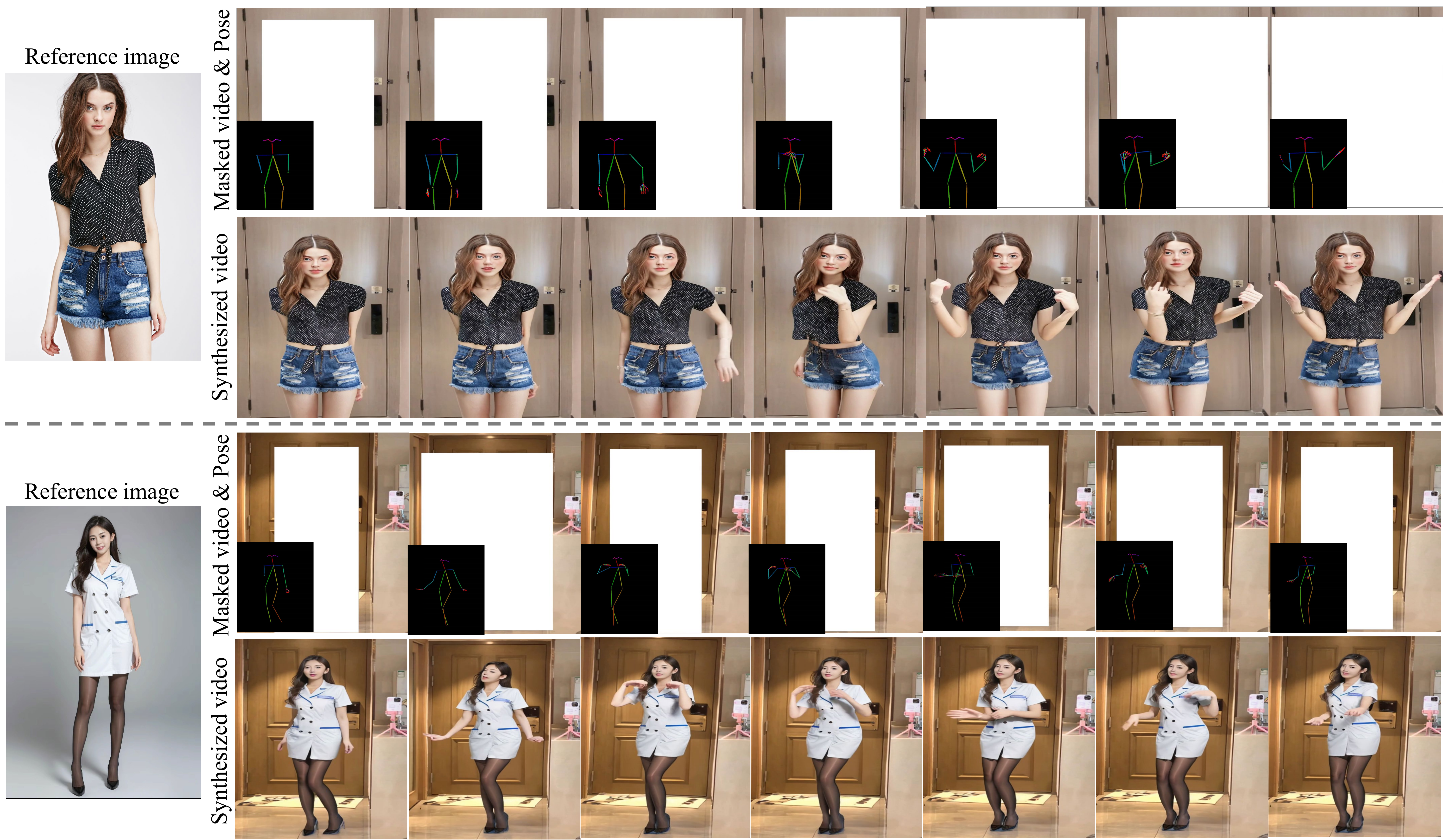}
    \vspace{-3mm}
    \caption{
        \textbf{{Qualitative results of character insertion}}. 
        The reference identity can be seamlessly inserted  into the masked video while showing the movements that follow the driven human poses.
    }
    \label{03_insertion}
    }
\end{figure*}

\vspace{0.5mm}
\noindent \textbf{Character insertion.}
\wx{
The task of character insertion involves inserting a new character into a coarse masked region of a video while maintaining the desired pose and appearance.
}
Regarding character insertion, we provide the visualizations in ~\cref{03_insertion}. From the results, we can observe that \method enables realistic video synthesis,
validating the capability of seamlessly integrating new characters into dynamic scenes. 
The generated videos effectively adhere to the pose sequence while ensuring the background remains undistorted.

\subsection{Ablation study}

\begin{figure*}[htbp]
{
        \centering
    \includegraphics[width=1.0\textwidth]{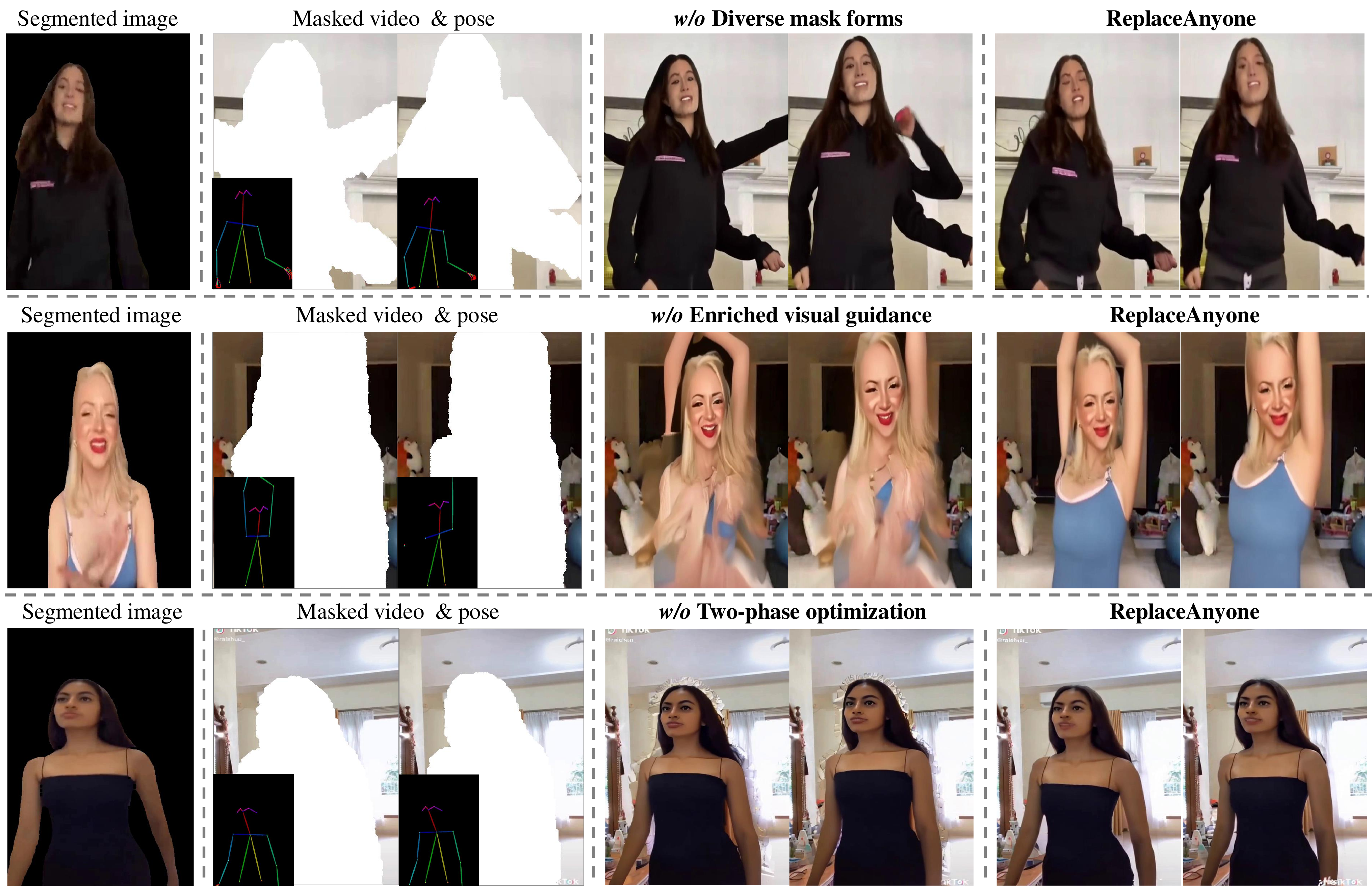}
    \vspace{-7mm}
    \caption{
        {\textbf{Ablation study} on the TikTok dataset}. 
        %
        The comparative results reveal the effectiveness and 
 vital role of each proposed component.
 The proposed \method can naturally and smoothly replace the person in the source video with the reference identity while maintaining the desired human poses.
    }
    \label{05_ablation}
    }
    \vspace{-3mm}
\end{figure*}

To achieve precise and localized control over human appearance and motion, 
we attempt to integrate image-conditioned pose-driven video generation and masked video inpainting tasks into a unified framework and introduce several novel components, including diverse mask forms, an enriched visual guidance mechanism, a hybrid inpainting encoder, and a two-phase training strategy.
These components collectively address the challenges of shape leakage, appearance alignment, background preservation, and optimization complexity.
We conduct ablation studies to assess the impact of different components of the proposed \method.
The results, as detailed in~\cref{tab:ablation_TikTok}, highlight the importance of each component in achieving optimal performance.
When each module is discarded, the performance will degrade to a certain extent.
Quantitatively, \method achieved a PSNR of 20.19, SSIM of 0.820, LPIPS of 0.187, and FVD of 183.66 on the TikTok dataset. These results demonstrate the high fidelity and temporal coherence of the generated videos. 
The framework effectively captured the visual characteristics of the reference character and integrated them seamlessly into the video, ensuring that the inserted character appeared natural and consistent with the background.
%
%
The additional visualizations in~\cref{05_ablation} also display the role of each component in generating visually appearing and temporally coherent videos.
For example, if the diverse mask forms  are not applied, the shape leakage problem may appear, causing an overfitting of the generated result onto the mask shape, such as the occurrence of four human hands.
We can also observe that the proposed two-phase optimization can promote harmony between the foreground and background boundary areas, and the enriched visual guidance mechanism helps to enhance the appearance alignment between the reference image and the synthesized video.
%

\begin{table}[t]
%
\caption{
\textbf{{Quantitative ablation studies}}  on  the TikTok dataset.
}
\vspace{-2mm}
\footnotesize
\renewcommand{\arraystretch}{1.2}
\setlength\tabcolsep{4.9pt}
\centering
\begin{tabular}{l|cccc}
\shline
\hspace{-1.5mm}Method          & PSNR$\uparrow$ & SSIM$\uparrow$ & LPIPS$\downarrow$  & FVD$\downarrow$ \\ \shline
\hspace{-1.5mm}\emph{w/o} Diverse mask forms 
 
  &    16.76     &       0.749      &          0.275                 &    405.08         \\
\hspace{-1.5mm}\emph{w/o}  Enriched visual guidance

&    18.08      &    0.766          &                    0.244       &      327.50       \\
\hspace{-1.5mm}\emph{w/o} Two-phase optimization 

&    17.09     &       0.754      &          0.286                 &      432.17       \\

\hspace{-1.5mm}{\textbf{\method}}
      & \textbf{20.19}             & \textbf{0.820}            & \textbf{0.187}                                          &\textbf{183.66}  \\

\shline
\end{tabular} 
\label{tab:ablation_TikTok}
\end{table}

\begin{table}[t]
\caption{
\wx{
\textbf{Ablation studies about the effect of hybrid inpainting encoder}  on  the TikTok dataset.
}
}
\vspace{-2mm}
\renewcommand{\arraystretch}{1.2}
\setlength\tabcolsep{3.1pt}
\centering
\begin{tabular}{l|cccc}
\shline
\hspace{-1.1mm}Method          & PSNR$\uparrow$ & SSIM$\uparrow$ & LPIPS$\downarrow$  & FVD$\downarrow$ \\ \shline
\hspace{-1.1mm}VAE encoder 

&    19.12     &       0.798      &          0.211                 &      251.33       \\
\hspace{-1.1mm}Learnable inpainting encoder 

&    18.77     &       0.769      &          0.242                 &      260.01       \\

\hspace{-1.1mm}{\textbf{Hybrid inpainting encoder (ours)}} 
      & \textbf{20.19}             & \textbf{0.820}            & \textbf{0.187}                                          &\textbf{183.66}  \\

\shline
\end{tabular} 
\label{tab:ablation_TikTok_inpainting_encoder}
\end{table}

\begin{figure}[t]
{
        \centering
    \vspace{0.6mm}\includegraphics[width=0.49\textwidth]{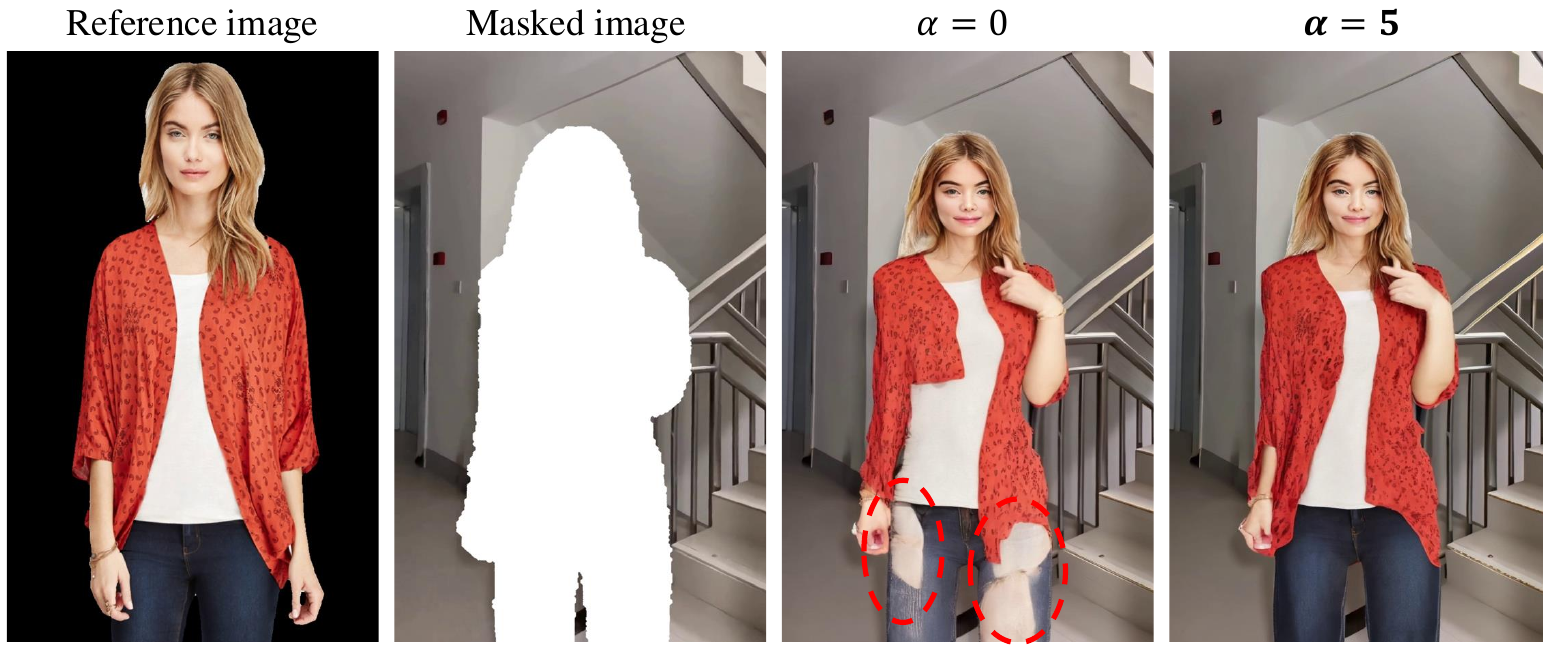}
    \vspace{-8mm}
    \caption{
    \wx{
        \textbf{Qualitative ablation studies on mask focused loss}.
        } Without the mask focused loss, the filling result may be difficult to
keep consistent with the appearance of the input reference
character.
    }
    \label{compare_loss}
    }
    \vspace{-2mm}
\end{figure}

\begin{figure*}[t]
{
        \centering
    \includegraphics[width=1.0\textwidth]{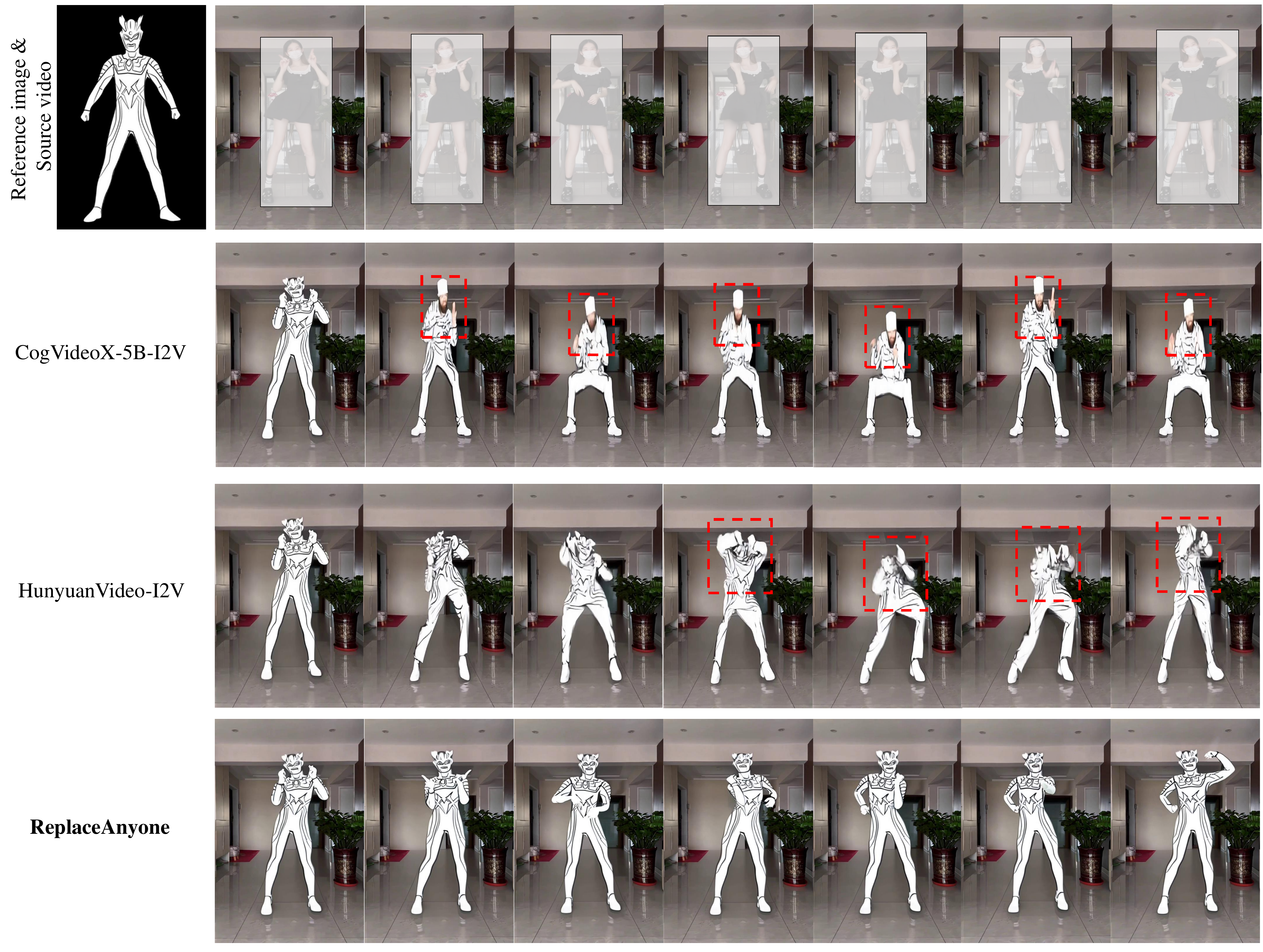}
    \vspace{-7mm}
    \caption{
        \wx{\textbf{Qualitative comparison with existing state-of-the-art image-to-video models}.
        We compare the proposed \method with CogvideoX-5B-I2V~\cite{yang2024cogvideox} and HunyuanVideo-I2V~\cite{kong2024hunyuanvideo}.
        }
    }
    \label{11_hunyuanvideo}
    }
    \vspace{-2mm}
\end{figure*}

\wx{
In~\cref{tab:ablation_TikTok_inpainting_encoder}, we show the performance of the hybrid inpainting encoder with a standard VAE encoder. 
The hybrid inpainting encoder improves performance compared to using a VAE encoder alone. This demonstrates the effectiveness of combining a complementary light-weight learnable encoder with the VAE encoder to enhance detail preservation ability.
In addition, we conduct a experiment, where a learnable light-weight inpainting encoder without VAE is used to inpaint videos, \ie, ``Learnable inpainting encoder" in ~\cref{tab:ablation_TikTok_inpainting_encoder}.
From the results, we can observe that ``Learnable inpainting encoder" underperforms the hybrid inpainting encoder.  We attribute this to the fact that the VAE encoder can help map the masked video to the denoised space, making it easier to learn, while the learnable inpainting encoder needs to learn to map to the denoised space, which is very difficult.
%
}

\wx{
In~\cref{compare_loss}, we compare the qualitative comparison results using the proposed mask focused loss and without the mask focused loss. It can be seen that without the addition of the mask focused loss, the filling result may be difficult to keep consistent with the appearance of the input reference character, and there will be a phenomenon of appearance misalignment. By introducing the mask focused loss, the attention on the area to be filled is enhanced during the training process, and the alignment effect is improved.
}

\begin{figure*}[t]
{
        \centering
    \includegraphics[width=1.0\textwidth]{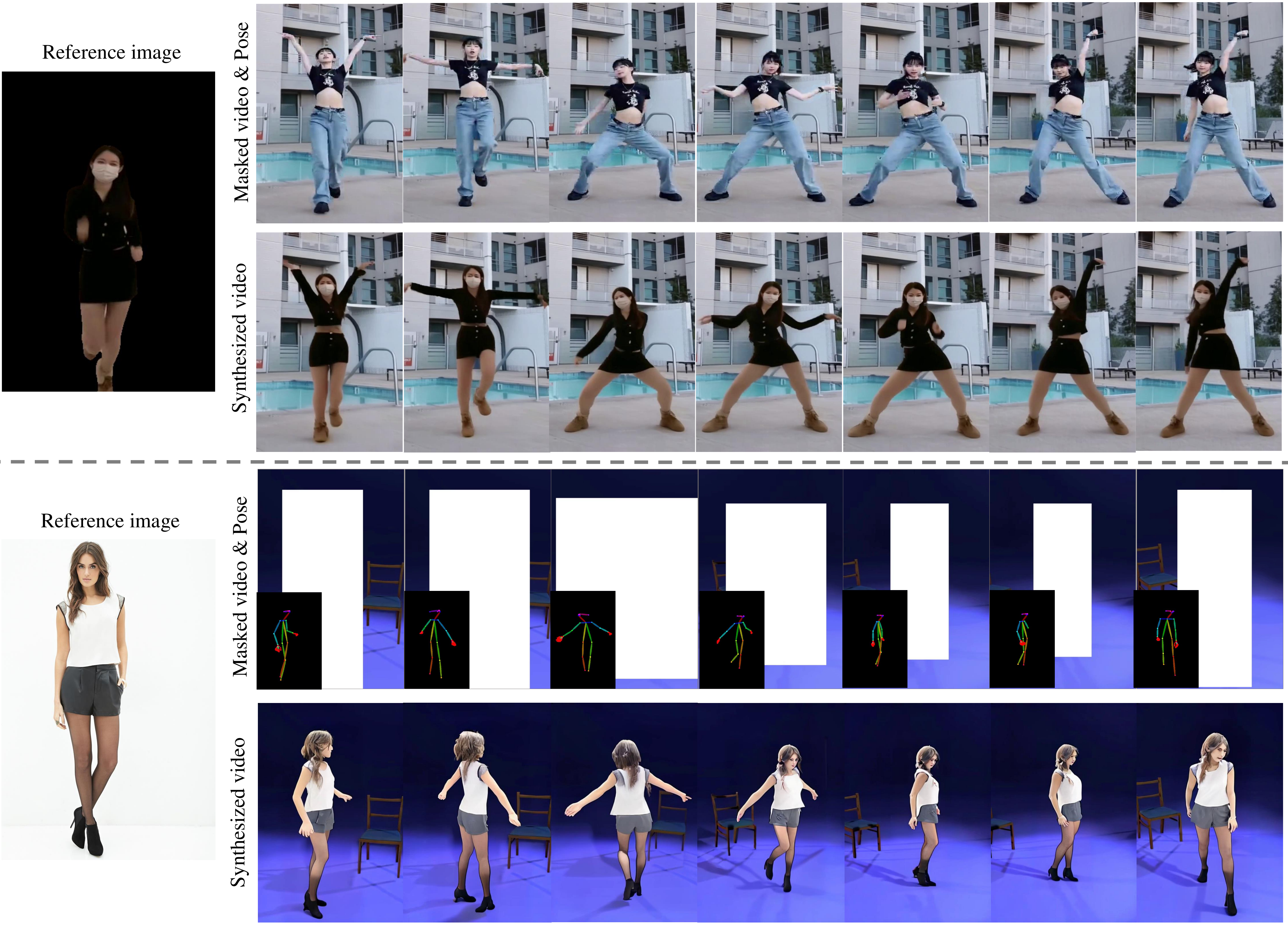}
    \vspace{-7mm}
    \caption{
        \wx{\textbf{Qualitative results with large background movement}.
        Our \method can be well generalized to large background motion scenes.
        }
    }
    \label{12_large_motion}
    }
    \vspace{-2mm}
\end{figure*}

\begin{figure*}[t]
{
        \centering
    \includegraphics[width=0.99\textwidth]{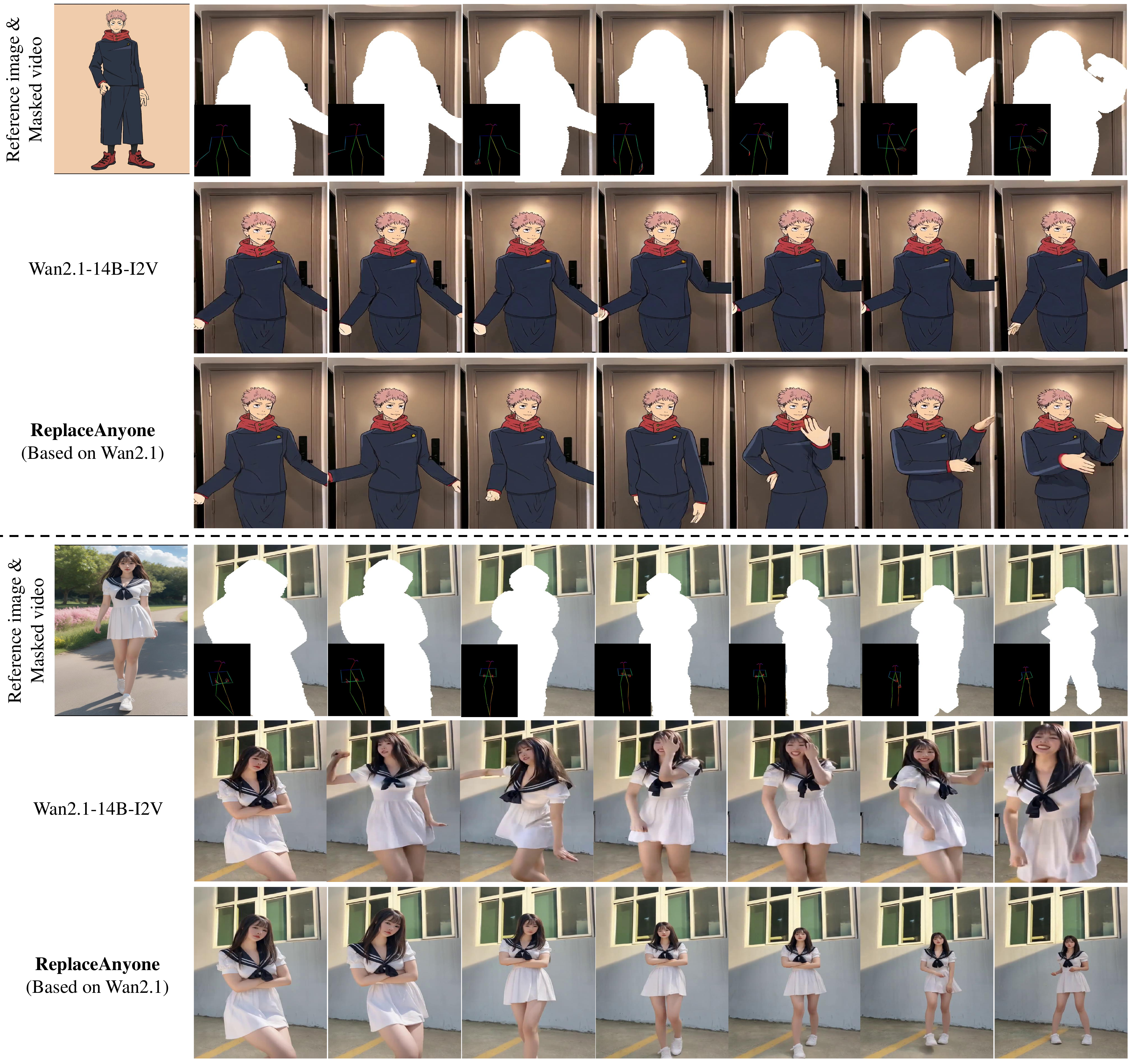}
    \vspace{-4mm}
    \caption{
        \wx{\textbf{Extending the proposed method to DiT-based Wan2.1 model}~\cite{wang2025wan}.
        The original Wan2.1 image-to-video model cannot maintain customized character poses and background motion, while our method can seamlessly insert characters into background videos with good temporal coherence.
        }
    }
    \label{13_wan}
    }
    \vspace{-2mm}
\end{figure*}
\wx{
In summary,
the success of \method in character insertion can be attributed to several key components of the framework:
1) Diverse mask forms: By using diverse mask forms, the model is able to prevent shape leakage and ensured that the inserted character did not exhibit artifacts or inconsistencies.
2) Enriched visual guidance: The enriched visual guidance mechanism extracts detailed features from the reference image, ensuring that the appearance of the inserted character aligned closely with the reference.
3) Hybrid inpainting encoder: The hybrid inpainting encoder preserves detailed background information, allowing the inserted character to blend harmoniously with the untouched background.
4) Two-phase optimization strategy: The two-phase training strategy simplifies the complexity of optimization.
These components collectively enabled \method to achieve seamless character insertion\&replacement, even in complex dynamic scenes.
}

\subsection{Comparison with image-to-video models}
\label{Comparison_with_image-to-video_models}

\wx{
Recent advancements in image-to-video (I2V) task~\cite{yang2024cogvideox,kong2024hunyuanvideo,zhang2023i2vgen} have achieved remarkable success, showing promising results in generating visually appearing and temporally coherent videos from a single input image. Notable methods such as CogVideoX-5B-I2V~\cite{yang2024cogvideox} and HunyuanVideo-I2V~\cite{kong2024hunyuanvideo} have demonstrated the ability to take an initial frame and generate subsequent frames, effectively creating temporally consistent video sequences. These models leverage the initial frame to establish the visual context and then extend it over time, maintaining temporal coherence.
Despite these advancements, in human-centric video generation task, maintaining consistent appearance and stable motion across frames remains a significant challenge. To evaluate the performance of existing I2V models in our task of localized human replacement and insertion, we conduct experiments where we provide the edited first frame (containing the replaced or inserted character) to these models and try to verify their ability to generate coherent video sequences.
As shown in~\cref{11_hunyuanvideo},
when provided with the edited first frame of a sketch character, CogVideoX-5B-I2V~\cite{yang2024cogvideox} struggles to preserve consistent appearance, particularly in facial regions. CogVideoX-5B-I2V exhibits noticeable changes in facial details across frames, leading to a disjointed visual experience. This inconsistency is primarily due to the model's difficulty in capturing the cross-domain fine-grained details required for stable appearance preservation.
In addition,
as observed in our experiments, HunyuanVideo-I2V~\cite{kong2024hunyuanvideo} exhibits video degradation over time, with the human body structure becoming unstable. 
It is extremely difficult to keep physically plausible motion for these models.
In contrast to these existing I2V models, our \method approach leverages pose guidance and enriched visual features to achieve seamless and consistent video generation. As displayed in~\cref{11_hunyuanvideo}, \method ensures that the generated video maintains the desired motion patterns and appearance consistency, indicating the effectiveness of the proposed method.
}

\subsection{Qualitative experiments in large motion scenes}

To further validate the effectiveness of the proposed \method framework, we conduct additional qualitative experiments focusing on large motion scenes. 
One of the primary challenges in large motion scenes is maintaining consistent appearance across frames.
The results are displayed in~\cref{12_large_motion}.  
From the results, we can see that \method demonstrates superior performance in maintaining consistent appearance.
These experiments highlight \method's ability to seamlessly insert or replace characters in videos with significant motion dynamics, which is a particularly challenging scenario for video generation and editing tasks, indicating the framework's robustness.

\subsection{Extension to DiT-based Wan2.1 model}

In previous research on controllable video generation~\cite{videocomposer,Animateanyone}, typical methods have predominantly relied on 3D-UNet architectures~\cite{modelscopet2v,tft2v} to generate videos. In this paper, to ensure a fair comparison with existing approaches, we also employ a 3D-UNet framework. With the advances of the DiT model~\cite{peebles2023scalable} in the image domain, there has been a growing interest in exploring DiT-based models for video generation~\cite{wang2025wan,kong2024hunyuanvideo,yang2024cogvideox}, leading to significant progress and improved video quality. To investigate the scalability and pluggability of our approach, we conduct additional experiments based on a state-of-the-art DiT-based model, namely Wan2.1~\cite{wang2025wan}.
Wan2.1 is able to achieve significant advancements
in generative capabilities through some dedicated strategies, including a novel
spatio-temporal variational autoencoder, scalable DiT training techniques, and the curation of
large-scale datasets.
In our experiment, the open-sourced Wan2.1-14B-I2V model is leveraged.
As shown in~\cref{13_wan}, we conduct comparisons where we feed the edited first frame generated by our \method to the Wan2.1 model and attempt to generate subsequent video frames and motion dynamics.
We can observe that
the original Wan2.1 image-to-video model can produce temporally consistent videos, but cannot display customized character poses and background motion since only a reference image and a textual prompt can't accurately depict the desired human motions. 
In contrast,
our \method can be integrated into the Wan2.1 model, producing high-quality videos with excellent temporal coherence and character fidelity, and can seamlessly insert characters into background videos with favorable temporal coherence and desired motions.
In addition, our method can not only achieve good fidelity for realistic characters, but also transfer well to some cartoon characters.
This demonstrates the scalability of our method and indicates that future enhancements can be achieved by building upon more advanced video models, potentially yielding even better results.
\section{Conclusion and limitations}

In this paper, we present a unified end-to-end framework named \method, allowing for localized human replacement and insertion within dynamic video scenes. 
Our methodological contributions, including diverse mask forms, an enriched visual guidance, a hybrid inpainting encoder, and a two-phase training strategy, substantiate the efficient integration of generated content.
Our framework addresses the significant challenges of achieving precise control over human motion and appearance while maintaining the authenticity and coherence of the dynamic scenes. 
Extensive experimental results demonstrate the validity of our framework over other baselines, paving the way for subsequent research.

\vspace{2mm}
\noindent{\textbf{Limitations.}
%
Despite the significant progress made by the proposed \method, there are still challenges that need to be addressed in future work. 
Firstly, handling inaccurate masks remains a challenge, as they can lead to visible artifacts and inconsistencies in the generated videos. Developing tolerance strategies for inaccurate masks and enhancing the robustness of the model to such errors is a promising direction for future research.
Secondly,
in some cases, the quality of facial and finger details in the generated videos is still limited by the base model, training data, and the pose estimation algorithms. 
Exploring advanced techniques for post-processing enhancement and incorporating specialized models for high-fidelity detail generation can further improve the realism of the generated content.

\section*{Acknowledgment}
This work is supported by the National Natural Science Foundation
of China under grants U22B2053 and 623B2039, and Alibaba Group through Alibaba Research Intern Program.

\vspace{-1mm}
{
\bibliographystyle{IEEEtran}
\bibliography{ref.bib}
}


\end{document}